\documentclass[manuscript,sigconf,nonacm]{acmart}
\usepackage{enumitem}
\usepackage{tabularx}
\usepackage{subcaption}
\usepackage{listings}
\usepackage{makecell}

\AtBeginDocument{%
  }

\setcopyright{acmlicensed}
\copyrightyear{2026}
\acmYear{2026}
\acmDOI{XXXXXXX.XXXXXXX}
\acmConference[KDD '26]{Proceedings of the 32nd ACM SIGKDD Conference on Knowledge Discovery and Data Mining}{August 9--13, 2026}{Jeju, Korea}
\acmISBN{979-8-4007-1454-2/2026/08}

\begin{document}

\title{ARCTraj: A Dataset and Benchmark of Human Reasoning Trajectories for Abstract Problem Solving}


\author{Sejin Kim}
\email{sejinkim@gist.ac.kr}
\affiliation{%
  \institution{GIST}
  \city{Gwangju}
  \country{Korea}
}

\author{Hayan Choi}
\email{hayan.choi@vtov.kr}
\affiliation{%
  \institution{VTOV}
  \city{Seoul}
  \country{Korea}
}

\author{Seokki Lee}
\email{sklee1103@gm.gist.ac.kr}
\affiliation{%
  \institution{GIST}
  \city{Gwangju}
  \country{Korea}
}

\author{Sundong Kim}
\email{sundong@gist.ac.kr}
\affiliation{%
  \institution{GIST}
  \city{Gwangju}
  \country{Korea}
}

\renewcommand{\shortauthors}{Kim et al.}

\begin{abstract}
We present ARCTraj, a dataset and methodological framework for modeling human reasoning through complex visual tasks in the Abstraction and Reasoning Corpus (ARC). 
While ARC has inspired extensive research on abstract reasoning, most existing approaches rely on static input-output supervision, which limits insight into how reasoning unfolds over time. 
ARCTraj addresses this gap by recording temporally ordered, object-level actions that capture how humans iteratively transform inputs into outputs, revealing intermediate reasoning steps that conventional datasets overlook. 
Collected via the O2ARC web interface, it contains around 10,000 trajectories annotated with task identifiers, timestamps, and success labels across 400 training tasks from the ARC-AGI-1 benchmark. 
It further defines a unified reasoning pipeline encompassing data collection, action abstraction, Markov decision process (MDP) formulation, and downstream learning, enabling integration with reinforcement learning, generative modeling, and sequence modeling methods such as PPO, World Models, GFlowNets, Diffusion agents, and Decision Transformers. 
Analyses of spatial selection, color attribution, and strategic convergence highlight the structure and diversity of human reasoning. 
Together, these contributions position ARCTraj as a structured and interpretable foundation for studying human-like reasoning, advancing explainability, alignment, and generalizable intelligence.
\end{abstract}




\maketitle

\section*{Acknowledgments}
As this work introduces ARCTraj as a complementary dataset for studying human strategies in ARC, we gratefully acknowledge the contributions of those who contributed to its creation. We thank the \textbf{developers of the O2ARC platform}, including Subin Kim, Prin Phunyaphibarn, Doyoon Song, Sanha Hwang, Hosung Lee, Suyeon Shim, Dohyun Ko, and Kyungmin Choi, for their continuous support during system development and maintenance. We are also grateful to the \textbf{organizers and participants of Happy ARC Day} for contributing valuable trajectories. Finally, we sincerely thank the \textbf{anonymous O2ARC users} whose interactions made the ARCTraj dataset possible. Their collective efforts enriched the trajectories and demonstrated the value of community-driven contributions to research on structured problem solving.

\section{Introduction}
\label{sec:intro}

Understanding and modeling how humans reason and solve problems, rather than reproducing only their final answers, remains a central challenge in AI~\citep{collins2024building}.
Such problem-solving involves conceptual abstraction, attention shifts, and the use of flexible strategies, yet these abilities remain difficult for machines to emulate~\citep{reisberg2013oxford, ho2022people}.
The Abstraction and Reasoning Corpus (ARC)~\citep{chollet2019ARC} was introduced to benchmark these capabilities through grid-based tasks where solvers must infer rules from a small set of input-output examples.

While the benchmark itself is well designed, most ARC research has focused on reproducing outputs rather than modeling reasoning.
Early program synthesis approaches~\citep{banburski2020dreaming, xu2023graphs, ouellette2024towards, rocha2025ilpar, ferre2025madil} struggled to infer solution strategies because the benchmarks provide limited guidance for program construction.
To address this, several studies have attempted to extract auxiliary cues from input grids or from intermediate program execution results; however, these approaches have remained task-specific and brittle.
More recently, LLM-based methods~\citep{tan2024llms, lei2025reasoning, pourcel2025soar} have leveraged prior knowledge to generate diverse candidate programs, showing improved flexibility but still facing the fundamental challenge of an enormous search space without explicit reasoning guidance.

In parallel, test-time training approaches~\citep{akyurek2025tttarc, franzen2024architect, li2025combining} aim to enable on-the-fly adaptation during inference; however, their learning signals still rely on static supervision rather than the evolving structure of reasoning.
As a result, existing ARC approaches (e.g., program synthesis, neuro-symbolic reasoning, and test-time learning) remain limited by their reliance on static supervision and their inability to capture the unfolding of human reasoning over time.

To address this limitation, we present \textbf{ARCTraj}, a large-scale dataset of human reasoning trajectories collected while solving ARC tasks.
Each trajectory captures a temporally ordered sequence of object-level actions (i.e., moving, rotating, and flipping objects) that transform an input grid into its correct output.
By capturing these multi-step trajectories, ARCTraj provides explicit reasoning supervision that connects human perception, abstraction, and strategy formation, offering a structured alternative to unconstrained program search.
These logs were collected via the O2ARC web interface~\citep{shim2024o2arc}, which is designed to support natural human interaction with ARC problems.
Each trajectory is annotated with metadata, including timestamps, task identifiers, and success labels, enabling both learning and analysis.

\begin{figure*}[htbp]
  \centering
    \includegraphics[width=0.8\textwidth]{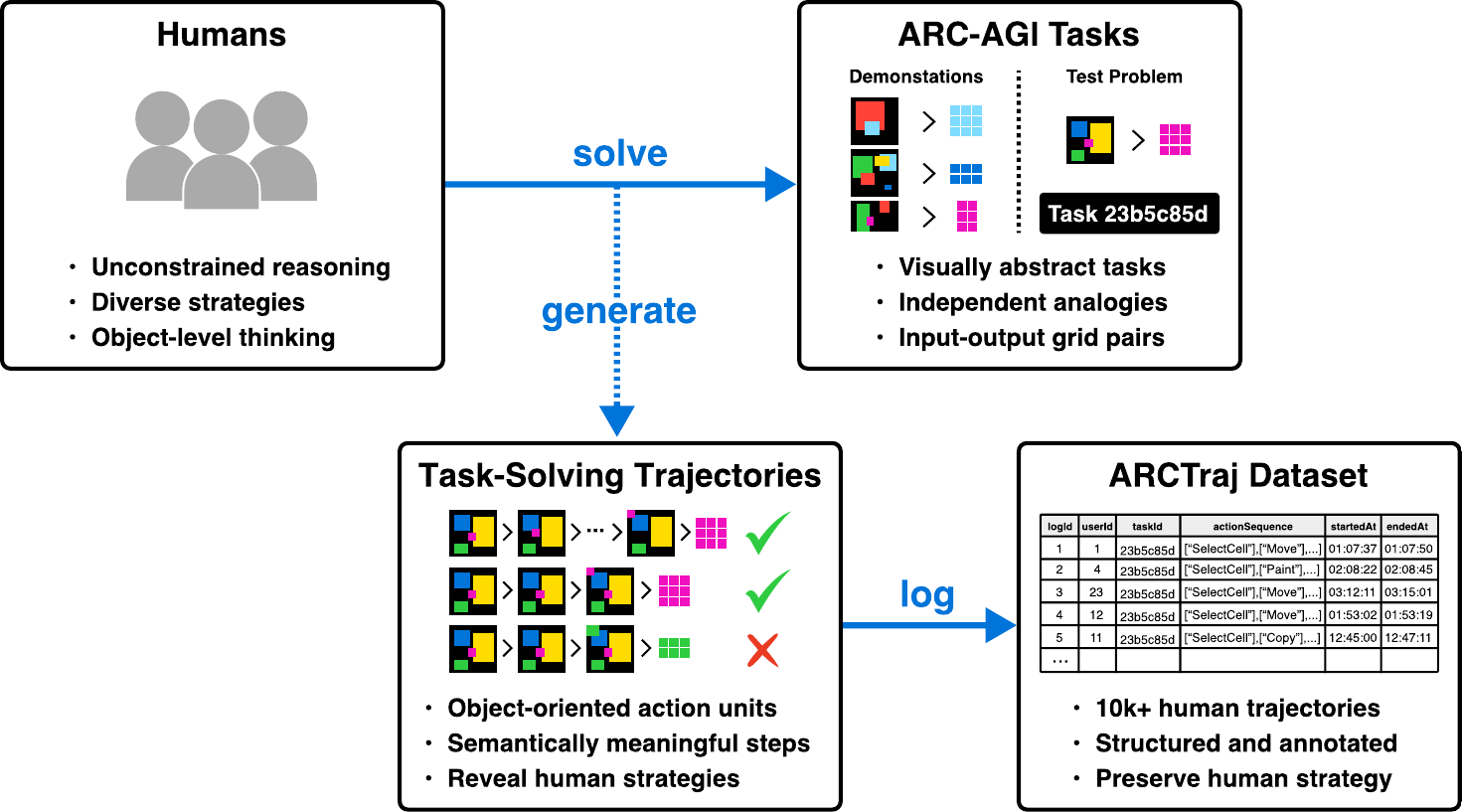}
    \caption{Overview of the ARCTraj data collection process. Users solve ARC tasks through the O2ARC platform by interacting with grid-based objects. Their actions are recorded step-by-step to form semantically rich, temporally ordered trajectories.}
    \Description{Overview of the ARCTraj data collection process. Users solve ARC tasks through the O2ARC platform by interacting with grid-based objects. Their actions are recorded step-by-step to form semantically rich, temporally ordered trajectories.}
    \label{fig:arctraj-generation}
\end{figure*}

Compared to existing human ARC datasets, ARCTraj offers several advantages.
Whereas H-ARC~\citep{legris2024harc} captures pixel-level edit logs and the ARC-Interactive-History-Dataset~\citep{neoneye2024arc-interactive} records low-level cell and operation sequences, ARCTraj provides object-centric actions with consistent formatting and semantic structure across all 400 ARC-AGI-1 training tasks.
This object-level abstraction reflects how humans group meaningful visual units and apply conceptual transformations rather than pixel-level edits.
Such abstraction preserves strategic intent and eliminates redundant operations, resulting in trajectories that more accurately represent human reasoning.

ARCTraj also defines a comprehensive reasoning pipeline that spans data collection, action abstraction, MDP formulation, and downstream learning.
This unified framework integrates naturally with reinforcement learning, generative modeling, and sequential reasoning methods that require temporally structured supervision.
Thus, ARCTraj serves not only as a dataset but also as a foundation for studying how reasoning behaviors can emerge from human trajectories.

Ultimately, ARCTraj shifts the focus of ARC research from reproducing outputs to modeling reasoning itself.
By learning from human trajectories, models can acquire structured strategies that reflect System 2 thinking~\citep{kim2024system} and generalize across domains, such as program synthesis, robotics, and data transformation.
We expect ARCTraj to promote research on transferable reasoning strategies beyond training tasks.

In this paper, we first introduce previous research (Sec.~\ref{sec:background}) and present the motivation and design of ARCTraj (Sec.~\ref{sec:dataset}).
We then formalize how ARC solvers learn from human trajectories through a unified reasoning framework (Sec.~\ref{sec:learning}).
Next, we present auxiliary knowledge extracted from ARCTraj, including selection, color, and intention cues (Sec.~\ref{sec:knowledge}).
Finally, we demonstrate their applications in ARC solvers and discuss broader implications for reasoning alignment between humans and AI (Sec.~\ref{sec:solvers}).

In summary, ARCTraj bridges a gap in the ARC domain by providing dynamic and interpretable records of human reasoning.
It supports both behavioral studies and the development of cognitively inspired models, serving as a versatile resource for research on abstraction, planning, and generalization.
\section{Related Work}
\label{sec:background}

\paragraph{Abstraction and Reasoning Corpus (ARC)}
ARC~\citep{chollet2019ARC}, also known as ARC-AGI, is a benchmark designed to test human-like generalization in abstract reasoning tasks. 
Each task consists of a few input–output grid pairs, requiring solvers to induce and apply conceptual transformations from limited examples. 
It has inspired research across multiple paradigms, including program synthesis~\citep{barke2024hysynth, bober2024neural, butt2024codeit}, neuro-symbolic reasoning~\citep{batorski2025nsa, liao2024compressarc, xu2023graphs}, and test-time training~\citep{akyurek2025tttarc, franzen2024architect, li2025combining}, many of which were featured in the ARC Prize 2024 Technical Report~\citep{chollet2024arcprize}. 
However, ARC remains static; it provides only input–output pairs, offering no insight into the intermediate reasoning that drives human problem-solving. 
As a result, most studies evaluate solution generalization rather than the reasoning process itself. 
Understanding this process is crucial for modeling human-like abstraction.
In this work, we use the term ARC to refer to the 400 training tasks from the ARC-AGI-1 benchmark, which form the foundation for all trajectories in ARCTraj.

\paragraph{Human Trajectory Datasets for ARC}
Several efforts have captured traces of human reasoning in ARC tasks. 
LARC~\citep{acquaviva2022communicating} collects natural language explanations, providing semantic insight but lacking action-level detail. 
Fast and Flexible~\citep{johnson2021fast} and H-ARC~\citep{legris2024harc} record pixel-level edits, while the ARC-Interactive-History-Dataset~\citep{neoneye2024arc-interactive} logs cell-level actions through the BrainGridGame interface. 
Although these datasets represent progress, their low-level granularity and inconsistent coverage make it challenging to extract structured reasoning or integrate data into learning frameworks. 
They reveal what humans edit, but not how strategic reasoning evolves. 
A higher-level, structured view is still missing. 
These gaps motivate ARCTraj, which captures object-level, temporally structured trajectories aligned across all ARC tasks to support both analysis and model training.
\begin{figure*}[htbp]
    \centering
    \includegraphics[width=0.8\linewidth]{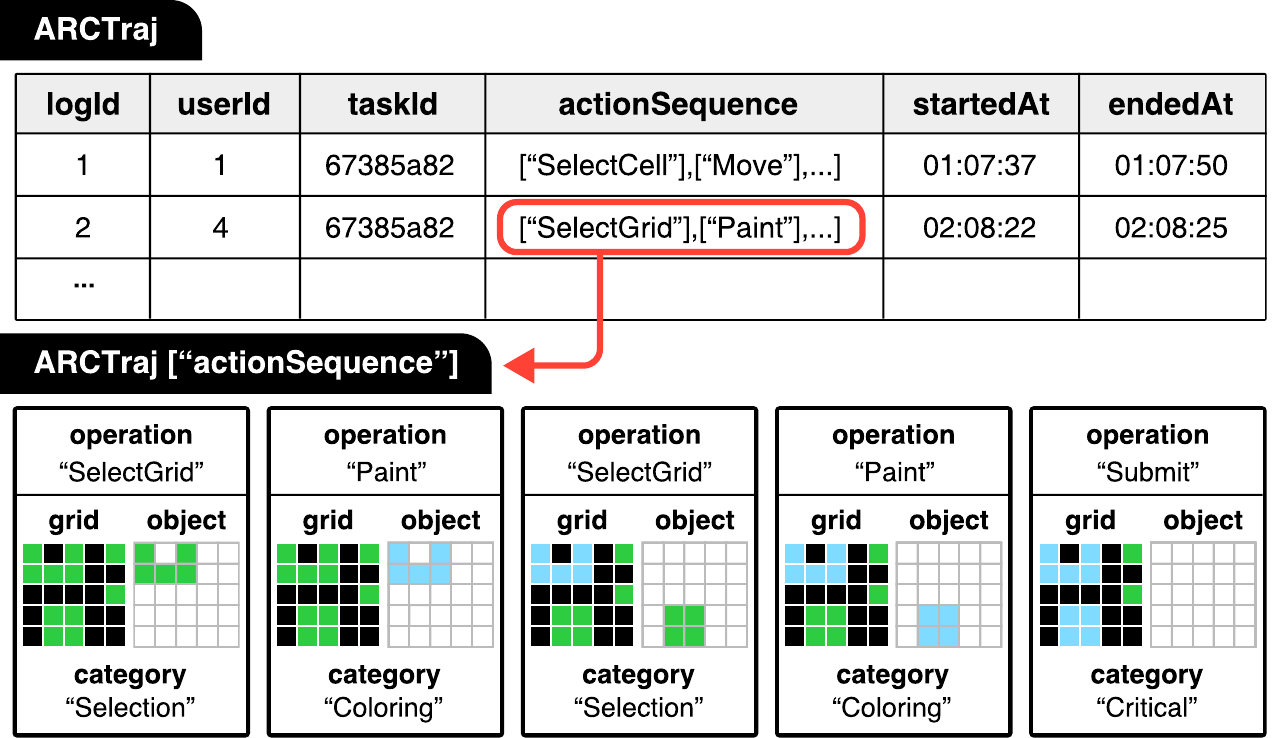}
    \caption{
    Example of a single trajectory log in ARCTraj.
    Each action in the ``\texttt{actionSequence}'' specifies its category and operation, along with the associated grid and object state, forming a structured state–action unit.
    }
    \Description{
    Example of a single trajectory log in ARCTraj.
    Each action in the ``\texttt{actionSequence}'' specifies its category and operation, along with the associated grid and object state, forming a structured state–action unit.
    }
    \label{fig:action-visual}
\end{figure*}

\section{ARCTraj Dataset}
\label{sec:dataset}

\subsection{Data Collection and Action Structure}
\label{sec:action-structure}

Fig.~\ref{fig:action-visual} illustrates the trajectory structure in ARCTraj and the components of each action.
Each log represents one human solving attempt and contains an ordered sequence of symbolic actions (\texttt{actionSequence}) recorded through the O2ARC interface~\citep{shim2024o2arc}.
Every action is represented as a triplet \(\langle \text{category}, \text{object}, \text{operation} \rangle\) associated with its grid state and timestamp.

In \texttt{actionSequence}, the \textbf{category} denotes the reasoning type involved (e.g., \textit{Selection}, \textit{Coloring}, \textit{Object-Oriented}, \textit{Clipboard}, \textit{Critical}), and the \textbf{object} indicates a perceptually meaningful region, contiguous colored cells automatically grouped by the interface but explicitly confirmed through user selection.
This user-driven segmentation ensures that objects reflect human perceptual grouping rather than heuristic clustering.
The \textbf{operation} specifies the action applied to the selected object (e.g., \texttt{Move}, \texttt{Paint}, \texttt{Flip}, or \texttt{Copy}), forming the decision component of the state–action pair.

These elements define an MDP-compatible format \(\langle s_t, a_t, s_{t+1} \rangle\), where \(s_t\) encodes the grid configuration and \(a_t\) represents the above triplet~\citep{lee2024arcle}.
Formally, the reward \(r_t \in \{1, 0\}\) indicates whether the current grid \(s_t\) matches the correct output.
By preserving order and object identity, ARCTraj captures evolving reasoning processes beyond static outcomes, enabling analysis of strategies such as hypothesis testing, selective attention, and object-level abstraction.

Each log also includes metadata, such as user ID, task ID, and timestamps, which enables the alignment and cross-user comparison of trajectories.
ARCTraj links every action to its grid context and semantic annotations, enabling replay, state comparison, and the extraction of reasoning sequences for downstream learning.

The dataset contains over 10,000 trajectories across all 400 ARC-AGI-1 training tasks, contributed by more than 300 participants.
Although the participant count is moderate, each user solved multiple tasks, producing diverse and deep trajectories that capture distinct problem-solving styles.
This design ensures both breadth across tasks and depth within each, supporting systematic analysis of human reasoning.
All participants consented to data collection under an Institutional Review Board (IRB)–approved protocol, and no personally identifiable information was stored.
\subsection{Comparative Dataset Statistics}
\label{sec:dataset-statistics}

Among existing human ARC datasets, the most representative one is H-ARC~\citep{legris2024harc}, which also records human solving processes on ARC tasks. 
However, the two datasets differ notably in both task coverage and the representation of actions. 
H-ARC collected trajectories from both the training and evaluation splits of ARC-AGI-1 and primarily logs pixel-level edits, where each action merges pixel selection and color change. 

In contrast, ARCTraj focuses on the 400 training tasks of ARC-AGI-1 and supports a broader range of object-related operations (\texttt{Move}, \texttt{Rotate}, \texttt{Flip}, \texttt{Copy}, and \texttt{Paste}). 
Because these operations are object-centric, ARCTraj records pixel selection as an explicit \texttt{Selection} action rather than embedding it within other edits. 

For fairness, all comparisons between ARCTraj and H-ARC in this section are limited to the 400 training tasks of ARC-AGI-1. 
All action-related metrics for ARCTraj (e.g., number of actions, ratio of object-related actions) include \texttt{Selection} steps by default. 
The values in parentheses exclude them for equivalence with H-ARC, which does not distinguish \texttt{Selection} from other actions.
This alignment ensures a consistent basis for measuring abstraction level and reasoning diversity across the two datasets.

\begin{table}[ht]
\centering
\caption{
Basic statistics of ARCTraj and H-ARC evaluated on the 400 training tasks of ARC-AGI-1. 
Both datasets are aligned to the same task split for consistent comparison.
}
\label{tab:arctraj_basic}
\begin{tabular}{@{}lrr@{}}
\toprule
Metric & ARCTraj & H-ARC \\
\midrule
Number of tasks & 400 & 400 \\
Number of participants & 100 & 783 \\
Number of trajectories & 10,672 & 7,916 \\
Number of unique visited states & 33,608 & 127,146 \\
Number of actions & 208,721 (84,123) & 241,697 \\
\bottomrule
\end{tabular}
\end{table}

ARCTraj comprises over 10,000 trajectories generated by about 100 participants who collectively solved the 400 training tasks of ARC-AGI-1. 
Although it involved fewer users than H-ARC, each participant solved multiple tasks, resulting in deeper and more complete reasoning trajectories. 
As shown in Table~\ref{tab:arctraj_basic}, ARCTraj records more trajectories despite fewer participants, indicating richer individual exploration and a wider variety of reasoning behaviors.

Because ARCTraj emphasizes object-level actions, it produces more abstract and efficient trajectories, leading to fewer unique visited states and fewer total action traces. 
The total action count includes \texttt{Selection} steps (208,721), which decreases to 84,123 when these steps are excluded. 
Overall, ARCTraj achieves extensive task coverage while maintaining depth within each task, providing a dense and diverse record of human problem-solving behavior.

\begin{table}[ht]
\centering
\caption{
Comparative abstraction statistics between ARCTraj and H-ARC. 
While both datasets share the same tasks, 
ARCTraj exhibits a broader exploration of reasoning strategies, 
a higher proportion of object-related actions, 
and a greater ratio of cross-trajectory grids, reflecting stronger object-level abstraction and consistent intermediate reasoning states.
}
\label{tab:arctraj_vs_harc}
\begin{tabular}{@{}lrr@{}}
\toprule
Metric & ARCTraj & H-ARC \\
\midrule
Avg. participants per task & 13.9 & 11.8 \\
Avg. trajectories per task & 25.5 & 19.8 \\
Ratio of object-related actions & 15.2\% (37.7\%) & 0.9\% \\
Ratio of cross-trajectory grids & 43.7\% & 11.4\% \\
\bottomrule
\end{tabular}
\end{table}

Beyond dataset scale, the abstraction level of reasoning behavior differs substantially between ARCTraj and H-ARC (Table~\ref{tab:arctraj_vs_harc}). 
Whereas H-ARC primarily logs low-level pixel edits, ARCTraj captures semantically meaningful, object-level manipulations that reduce redundancy and reveal compositional reasoning patterns aligned with human intuition. 
These higher-level representations make trajectories easier to interpret and model, providing a stronger basis for analyzing strategic reasoning and developing models that emulate human problem-solving behavior.

\begin{itemize}[leftmargin=*]
    \item \textbf{Broader participant diversity and strategic coverage.}  
    ARCTraj includes slightly more participants and trajectories per task than H-ARC, offering broader coverage of diverse human reasoning strategies.
    This variety captures differences in planning depth, exploration style, and strategic decision-making across solvers.

    \item \textbf{Higher-level reasoning and abstraction.}  
    The large proportion of object-related actions indicates that ARCTraj captures reasoning at a higher semantic level. 
    By emphasizing object-centric manipulation over pixel-level editing, it produces trajectories that mirror human conceptual understanding and provide supervision for model learning.

    \item \textbf{Convergent intermediate states.}  
    ARCTraj exhibits a higher ratio of cross-trajectory grids, suggesting that participants frequently converge on similar intermediate states despite employing distinct strategies. 
    Such convergence implies shared cognitive structures and stable reasoning pathways that support analyses of intention alignment and strategy generalization.
\end{itemize}

Taken together, these advantages establish ARCTraj as a valuable resource for research centered on reasoning.
Its structured, object-centric trajectories provide direct supervision signals for models that learn to imitate or infer human-like reasoning, as explored in the subsequent analyses and applications.
\section{Learning from ARCTraj}
\label{sec:learning}

\subsection{Formalizing ARC Solvers with ARCTraj}

To understand how ARCTraj contributes to solving ARC tasks, we formalize the ARC objective as a few-shot reasoning problem.
Each task provides $K$ demonstration pairs:
\[
\mathcal{D}_{\text{demo}} = \left\{\big(x_k^{\text{demo}}, y_k^{\text{demo}}\big)\right\}_{k=1}^K,
\]
where \(x_k^{\text{demo}}\) and \(y_k^{\text{demo}}\) denote input and output grids illustrating the transformation pattern.
The goal is to predict the output \(\hat{y}^{\text{test}}\) for a new input \(x^{\text{test}}\) using:
\[
f_\theta : (\mathcal{D}_{\text{demo}}, x^{\text{test}} [, \mathcal{A}]) \mapsto \hat{y}^{\text{test}},
\]
or equivalently,
\[
\hat{y}^{\text{test}} = \arg\max_{y'} P\big(y' \mid \mathcal{D}_{\text{demo}}, x^{\text{test}} [, \mathcal{A}]\big),
\]
where \(y'\) is a candidate output grid.

Unlike conventional supervised learning, ARC requires solvers to infer abstract rules from a few examples rather than memorizing mappings.
Thus, \(f_\theta\) must induce compositional concepts that explain unseen transformations, framing ARC as a meta-reasoning problem that demands generalization beyond training examples.

In this formulation, \(f_\theta\) takes the demonstrations and test input as inputs, optionally conditioned on auxiliary knowledge \(\mathcal{A}\).
Here, \(\mathcal{A}\) denotes \textit{auxiliary knowledge} that guides solvers toward human-like reasoning.
\(\mathcal{A}\) can include external cues such as visual priors, symbolic hints, or reasoning trajectories, among which ARCTraj offers a concrete instance capturing human action sequences.
While \(\mathcal{D}_{\text{demo}}\) provides static supervision, ARCTraj records intermediate steps that support compositional learning.

Its information about attention, color usage, and multi-step strategies offers structured supervision that aligns models with human reasoning.
This auxiliary knowledge enhances interpretability and robustness across diverse ARC tasks, defining ARC solving as conditional reasoning guided by general external knowledge, with ARCTraj as one realization of it.
The next section introduces specific forms of \(\mathcal{A}\), including selection, color, and intention cues derived from ARCTraj.

\begin{figure*}[htbp]
    \centering
    \includegraphics[width=\linewidth]{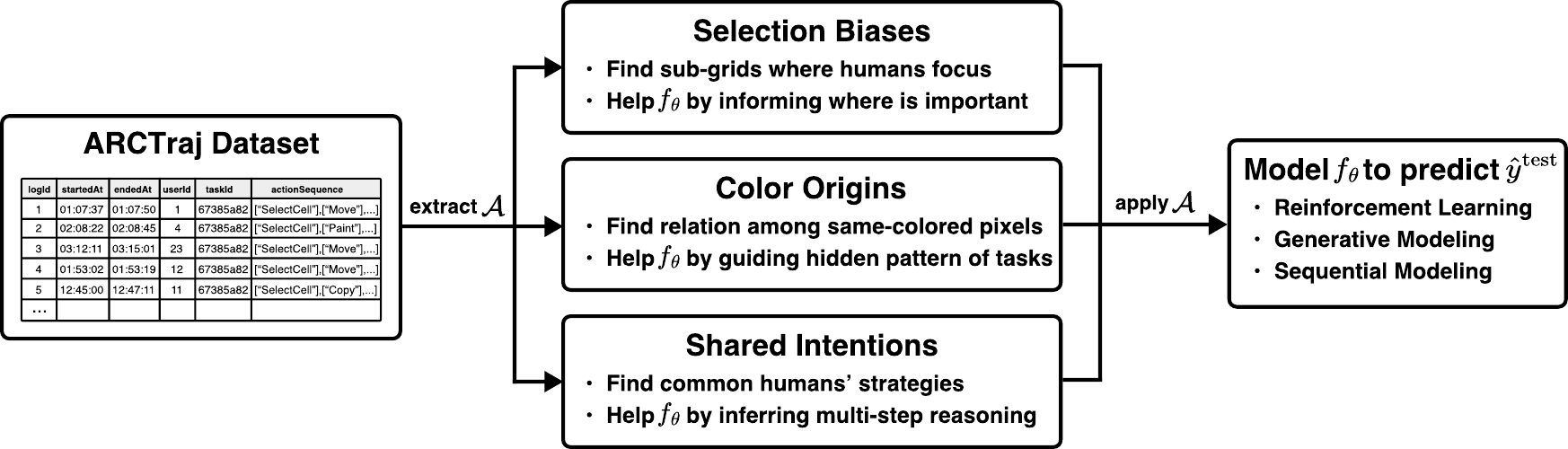}
    \caption{
    Overview of how ARCTraj analyses inform ARC solving.
    The ARC solver \( f_\theta \) predicts \(\hat{y}^{\text{test}}\) given \(x^{\text{test}}\), demonstration examples \(\mathcal{D}_{\text{demo}}\), and auxiliary knowledge \(\mathcal{A}\) derived from human trajectories.
    ARCTraj provides three structured components of \(\mathcal{A}\), (1) selection biases, (2) color origins, and (3) shared intentions, that capture different stages of human reasoning and can be integrated into model learning.
    }
    \Description{
    Overview of how ARCTraj analyses inform ARC solving.
    The ARC solver \( f_\theta \) predicts \(\hat{y}^{\text{test}}\) given \(x^{\text{test}}\), demonstration examples \(\mathcal{D}_{\text{demo}}\), and auxiliary knowledge \(\mathcal{A}\) derived from human trajectories.
    ARCTraj provides three structured components of \(\mathcal{A}\), (1) selection biases, (2) color origins, and (3) shared intentions, that capture different stages of human reasoning and can be integrated into model learning.
    }
    \label{fig:analysis_with_trajectory}
\end{figure*}

\subsection{Learning Paradigms with ARCTraj}

Fig.~\ref{fig:analysis_with_trajectory} illustrates how insights from ARCTraj reshape the learning perspective of ARC solving.
Rather than treating human trajectories as passive data, ARCTraj reframes them as structured supervision, revealing how humans perceive, infer, and act across reasoning stages.
The resulting auxiliary knowledge captures three complementary dimensions of cognition: \textit{where} attention is directed, \textit{what} information is abstracted, and \textit{how} strategies are organized.
These cues translate human reasoning patterns into actionable signals that guide model training, offering a new foundation for human-aligned learning.
ARCTraj bridges symbolic reasoning and empirical data, providing an interpretable scaffold through which models internalize human-like reasoning principles beyond statistical regularities.

Existing ARC solvers, though diverse in design, have yet to fully incorporate such structured reasoning cues.
Reinforcement learning methods exploit trajectory data for exploration and reward shaping, but often overlook the semantic structure behind human actions.
Generative models such as diffusion and GFlowNet solvers produce diverse solutions but lack interpretability grounded in perceptual or conceptual regularities.
Sequential models, including decision transformers and imitation-based systems, reproduce action sequences without modeling the intentions that guide them.
Most current solvers thus operate at the behavioral level of imitation rather than the cognitive level of reasoning abstraction, excelling at pattern reproduction but struggling to adapt when task structures change, highlighting the need for more cognitively grounded learning.

ARCTraj therefore suggests a shift in learning paradigm: from replicating task outcomes to modeling the reasoning process itself.
By integrating auxiliary knowledge (e.g., selection biases, color origins, and shared intentions), future solvers can achieve explainable, generalizable reasoning aligned with human cognition.
Such integration enables models not only to improve task accuracy but also to display interpretable intermediate reasoning that reveals the rationale behind their transformations.
ARCTraj is not merely a dataset but a framework that connects human cognitive analysis and machine learning design, paving the way for interpretable, human-centered ARC solvers.

\section{Auxiliary Knowledge from ARCTraj}
\label{sec:knowledge}

ARCTraj is not only a corpus of human trajectories but also a source of auxiliary knowledge about how people reason, explore, and strategize in solving ARC tasks.
This section distills that knowledge to reveal cognitive regularities in visual reasoning and cross-participant behavior.
Using quantitative and qualitative analyses, we examine how humans focus on relevant regions, infer transformation rules, and organize multi-step strategies to reach correct outputs with adaptive flexibility.

To structure this investigation, we frame three research questions (RQs) that correspond to the main stages of human problem solving, from perceptual attention to hypothesis formation and strategic abstraction~\citep{reisberg2013oxford, ho2022people}.
Each RQ highlights a distinct aspect of reasoning behavior, together forming a coherent picture of how humans decompose, transform, and recombine patterns in ARC task solving, providing insights that bridge human cognitive processes and AI reasoning frameworks~\citep{collins2024building}.

\begin{enumerate}[leftmargin=*, label=RQ~\arabic*.]
    \item \textbf{Selection Biases: ``Where'' do humans focus attention during problem solving? (Sec.~\ref{sec:selection_biases})}
    We analyze spatial and object-level \textit{selection biases} to identify which regions humans interact with most, how the number of objects relates to trajectory length, and whether attention patterns indicate perceptual or conceptual salience.

    \item \textbf{Color Origins: ``What'' patterns or information do humans infer when generating new outputs? (Sec.~\ref{sec:color_origin})}
    We investigate the \textit{origin of colors} in human-created test outputs to understand how participants infer generative rules and how these color cues reflect inductive biases relevant to model design.

    \item \textbf{Shared Intentions: ``How'' do humans construct and generalize multi-step reasoning strategies? (Sec.~\ref{sec:shared_intention})}
    We examine variations and convergences in solution trajectories across participants to reveal abstracted reasoning intentions and shared structural pathways, moving beyond trajectory clustering toward higher-level \textit{intention grouping} and convergence analysis.
\end{enumerate}

\subsection{Biases in Human Grid Selections}
\label{sec:selection_biases}

To address RQ1, we examine whether humans show systematic selection biases when interacting with ARC grids.
In ARCTraj, each selection action can involve a single pixel, a user-defined region, or an object-level selection that includes multiple pixels.
To unify these cases, we compute the bounding box for each selection, defined as the smallest axis-aligned rectangle enclosing all selected pixels.
We then analyze the distributions of bounding box height, width, and area to describe selection scale and shape across the dataset.

\begin{figure}[htbp]
    \centering
    \begin{subfigure}[b]{0.50\linewidth}
        \centering
        \includegraphics[width=\linewidth]{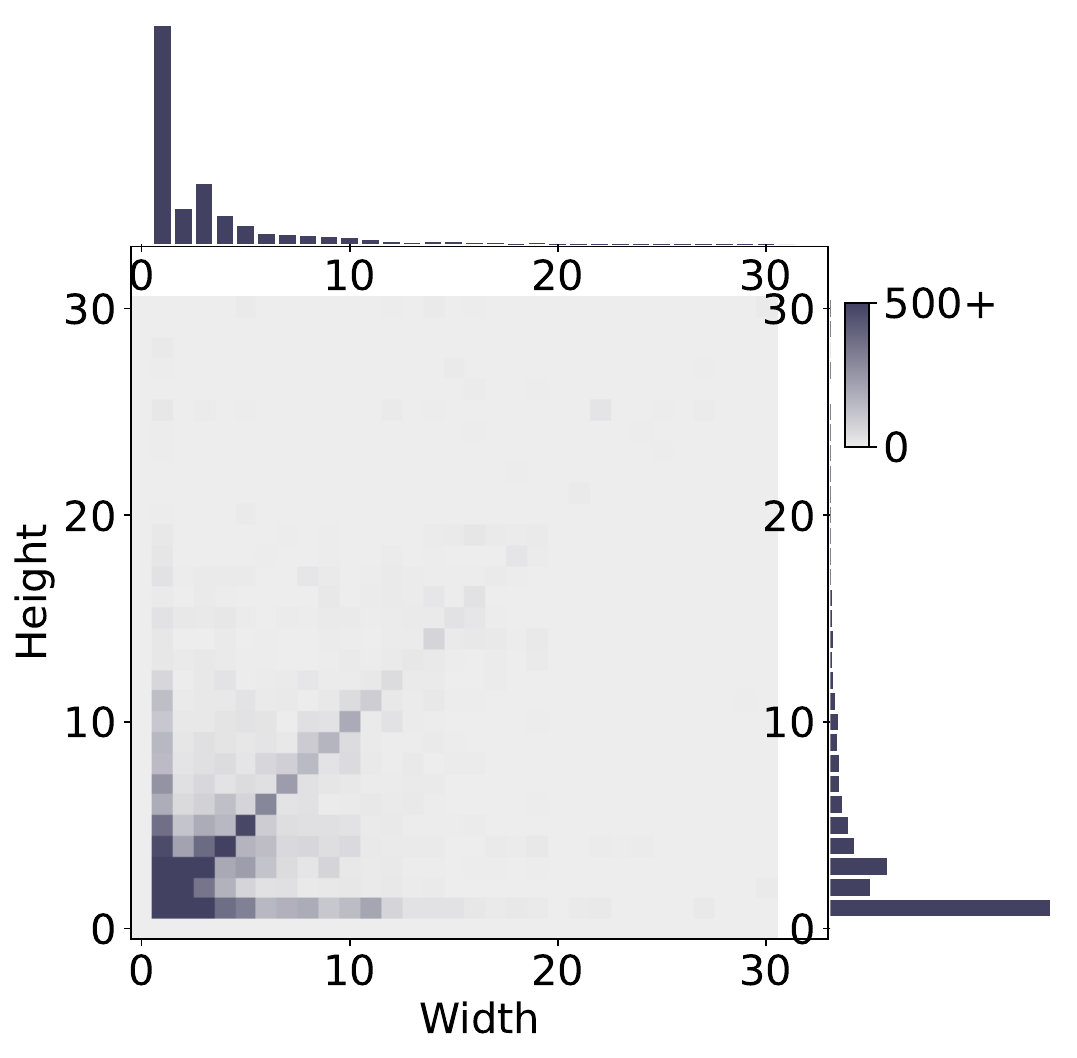}
        \Description{Joint distribution of selection height and width.}
        \label{fig:selection-heatmap}
    \end{subfigure}
    \hfill
    \begin{subfigure}[b]{0.48\linewidth}
        \centering
        \includegraphics[width=\linewidth]{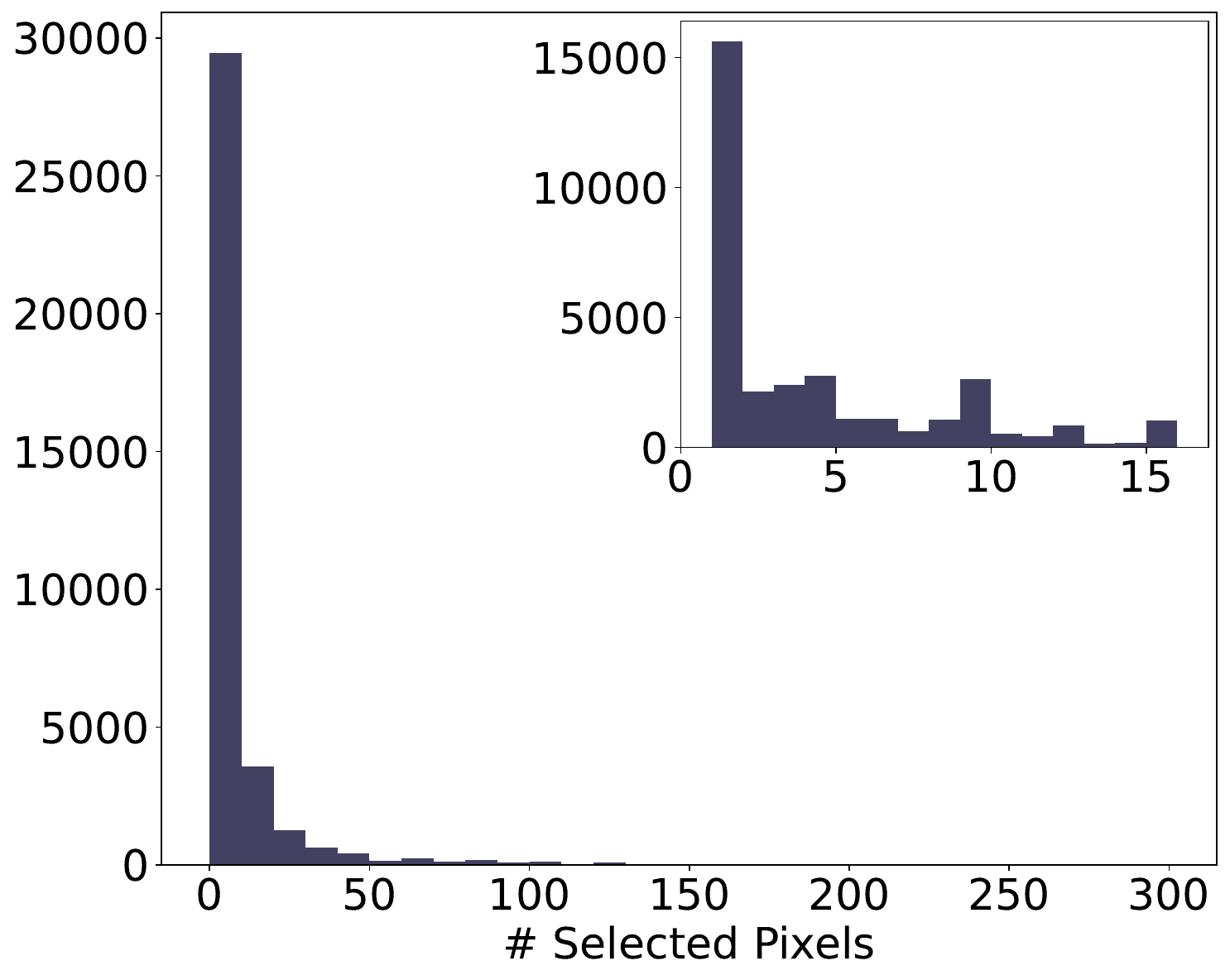}
        \Description{Histogram of the number of selected pixels.}
        \label{fig:selection-area-histogram}
    \end{subfigure}
    
    \caption{
    Distributions of human selection behavior in ARC tasks. 
    Left: Selections are concentrated in compact shapes ranging from $1\times1$ to $3\times3$, with square- and bar-shaped regions dominating. 
    Right: Most selections cover fewer than 20 pixels, reflecting a preference for local and perceptually salient regions.
    }
    \Description{
    Distributions of human selection behavior in ARC tasks. 
    Left: Selections are concentrated in compact shapes ranging from $1\times1$ to $3\times3$, with square- and bar-shaped regions dominating. 
    Right: Most selections cover fewer than 20 pixels, reflecting a preference for local and perceptually salient regions.
    }
    \label{fig:selection-behavior-combined}
\end{figure}

As shown in Fig.~\ref{fig:selection-behavior-combined}, we observe three main tendencies: (i) selected areas are mostly small (under $3 \times 3$), (ii) square-shaped selections ($n \times n$) dominate, and (iii) bar-shaped selections ($n \times 1$ or $1 \times m$) also appear frequently.
These patterns indicate a consistent preference for local reasoning and perceptually regular shapes.
The left panel shows the joint distribution of selection height and width, which is densely concentrated in the $1 \times 1$ to $3 \times 3$ range. A diagonal ridge reflects a square-shape bias, and off-diagonal clusters indicate bar-shaped regions.
The right panel shows the distribution of selected pixel counts, with most selections involving fewer than 16 pixels.
Peaks near square numbers (1, 4, 9, 16) further confirm humans’ bias toward compact, symmetric regions.

\textbf{Research Direction 1(a): Temporal Dynamics of Selection Behavior.}
Future work could explore how selection size and shape evolve.
Do humans begin with small exploratory selections and later expand them after identifying transformation patterns, or do they start globally and then focus on details?
Temporal analyses could reveal how attention shifts during problem solving and inspire phase-based attention models in AI.

\textbf{Research Direction 1(b): Perceptual Features and Selection Probability.}
Another direction is to test whether perceptual cues such as color contrast, isolation, or proximity to boundaries affect selection likelihood.
This can be studied through controlled manipulation of grid layouts and saliency.
Modeling this relationship could support predictive attention models and human-AI collaborative reasoning systems.

\subsection{Color Source Attribution in Test Outputs}
\label{sec:color_origin}

To address RQ2, we examine the sources of the colors used in the test output grids and how they relate to human color selection strategies.
Color plays a central role in ARC tasks, yet understanding how humans choose and transfer colors presents distinct challenges for trajectory-based analysis.
Unlike spatial selections or object manipulations, color decisions often occur implicitly, revealing how humans integrate perception and reasoning in non-verbal ways.

An examination of the ARC-AGI-1 training set and its corresponding trajectories reveals that output colors typically originate from a limited number of sources.
Among the 400 training tasks, 266 can be solved using only colors from the test input grid, while 134 require colors drawn from both the test input and the example output grids.
Notably, no task requires colors that appear exclusively in the example inputs, even when considering all possible sources (test input, example output, and example input).
This pattern indicates a deliberate design constraint that restricts potential color sources to the test inputs and, in some cases, the example outputs, thereby reducing task ambiguity and simplifying color-based reasoning.
It also implies that color selection is not random or heuristic, but instead follows consistent patterns shaped by task structure and perceptual cues.

While ARCTraj does not explicitly record where users obtained their colors, the observed color-selection patterns closely align with these potential sources.
Participants consistently use colors from the test inputs or example outputs, even without explicit color-sampling tools.
This suggests that humans implicitly perform color source attribution when reasoning about transformations, mentally tracking how colors relate across different grid examples and maintaining internal mappings between corresponding regions or objects.

\begin{table}[htbp]
    \centering
    \caption{Source of colorsets in ARC tasks. In 66.5\% of tasks, required colors appear in the test input grids only. The remaining 33.5\% require colors from both the test input and example output grids. No task requires colors exclusive to the example input.}
    \begin{tabular}{@{}lcc@{}}
        \toprule
        \textbf{Source of Colorset} & \textbf{\# of Tasks} & \textbf{\%} \\
        \midrule
        Test Input Grids Only & 266 & 66.5 \\
        + Example Output Grids (added) & 134 & 33.5 \\
        + Example Input Grids (added) & 0 & 0.0 \\
        \bottomrule
    \end{tabular}
    \label{tab:color-sources}
\end{table}

\textbf{Research Direction 2(a): Trajectory Logging with Color Origin Tracking.}
Future research could introduce trajectory-recording interfaces that explicitly log color-origin information.
By extending current DSLs with operators such as \texttt{sample\_color(grid, x, y)} or \texttt{apply\_color\_transformation(rule)}, users could directly sample and apply colors across grids. 
At the same time, the system records their relational origins.
This would enable detailed documentation of how humans establish color correspondences, producing richer datasets for evaluating model alignment with human color reasoning.
Such interfaces would bridge perceptual sampling and symbolic reasoning, providing more accurate supervision for color-based transformations and enabling models to learn transferable color-reasoning patterns.

\textbf{Research Direction 2(b): Generalized Origin Tracking.}
Beyond color, similar origin tracking can be applied to other task elements.
Objects, shapes, spatial configurations, and transformation rules may also be derived from specific examples within the ARC tasks.
Developing generalized frameworks for origin tracking would enable the analysis of how humans extract and reuse information across examples and tasks.
For instance, when constructing new grid structures, do humans reference example outputs, or when selecting specific object sizes, are they guided by test inputs?
This multidimensional analysis would deepen our understanding of cross-example analogical reasoning, revealing how humans anchor different aspects of the solution process and informing the design of models that replicate such structured reasoning.
\subsection{Shared Intentions and Strategy Patterns}
\label{sec:shared_intention}

To address RQ3, we analyze how humans construct and generalize multi-step reasoning strategies when solving ARC tasks.
Rather than comparing entire trajectories, we focus on mid-sequence decisions that reflect shared intentions, specifically choices about which region to act on and how to transform it.
This perspective captures human reasoning at an intermediate level between low-level operations and full solution paths, highlighting how solvers plan and adapt their actions dynamically.

In ARCTraj, each trajectory consists of alternating focus and transformation steps.
A focus action highlights a rectangular grid region, which may represent a complete object, a fragment, or a perceptually meaningful area.
After one or more focus steps, the user performs a transformation such as moving, coloring, deleting, or copying the selected region.
Human solvers do not always alternate strictly between focusing and acting; they often explore several areas sequentially to inspect structures or compare subgoals before committing to a transformation.
For example, a participant might highlight multiple red squares before recoloring one to match a blue pattern, reflecting a search-and-align reasoning process.

\begin{table}[htbp]
    \centering
    \captionof{table}{
    Distribution of selection actions preceding each operation. 
    Most operations are preceded by one to four selections, indicating that users tend to engage in short exploratory phases before executing a concrete transformation.
    }
    \begin{tabular}{@{}rrrr@{}}
        \toprule
        \textbf{Length} & \textbf{Count} & \textbf{\%} & \textbf{Cum. \%} \\
        \midrule
        1 & 23{,}632 & 63.7 & 63.7 \\
        2 & 5{,}451 & 14.7 & 78.4 \\
        3 & 3{,}343 & 9.0 & 87.4 \\
        4 & 1{,}379 & 3.7 & 91.1 \\
        $\vdots$ & $\vdots$ & $\vdots$ & $\vdots$ \\
        386 & 1 & 0.0 & 100.0 \\
        \bottomrule
    \end{tabular}
    \label{tab:length_distribution}
\end{table}

\begin{figure}[htbp]
    \centering
    \setlength{\belowcaptionskip}{-15pt}
    \includegraphics[width=\linewidth]{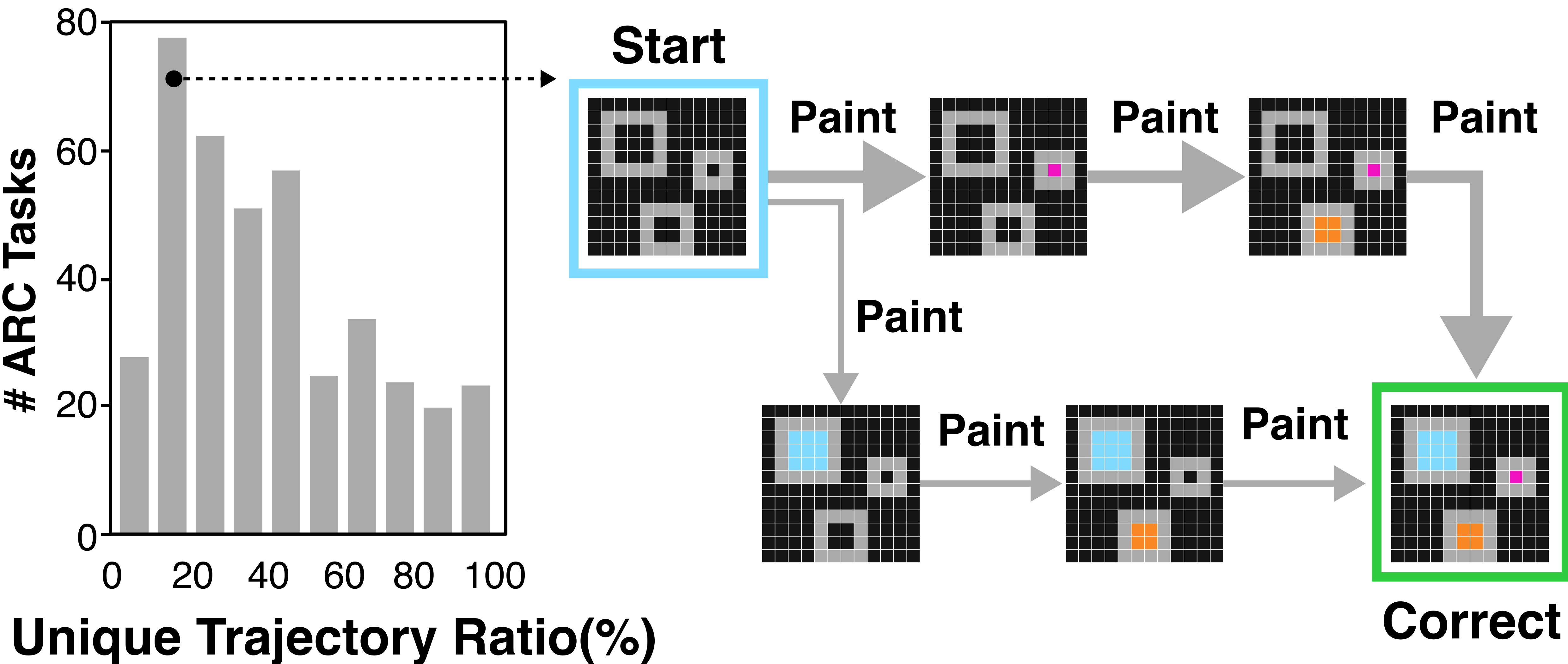}
    \caption{
    Uniqueness analysis of human reasoning trajectories.
    According to the left panel, most ARC tasks show low unique trajectory ratios, indicating that human solvers often converge on similar reasoning paths.
    The right panel shows a representative low-uniqueness task (\href{https://o2arc.com/task/c0f76784}{Task~c0f76784}; see Appendix~\ref{appendix:arc_task}), where overlapping solution routes appear in the state-space graph.
    }
    \Description{
    Uniqueness analysis of human reasoning trajectories.
    Most ARC tasks show low unique trajectory ratios, indicating that human solvers often converge on similar reasoning paths.
    The panel shows a representative low-uniqueness task (Task~c0f76784; see Appendix~\ref{appendix:arc_task}), where overlapping solution routes appear in the state-space graph.
    }
    \label{fig:compression_histogram}
\end{figure}

Our analysis shows that 63.7\% of operations are preceded by a single selection, and over 90\% occur within four selections (Table~\ref{tab:length_distribution}).
This indicates that humans typically perform a few attentional shifts before committing to a concrete transformation.
The short interval between selections and operations reflects rapid exploratory reasoning, as humans briefly scan local contexts before making targeted decisions.

To capture mid-level convergence, we define a shared intention as a pairing of a selection region and an operation type that recurs across participants solving the same task.
For example, if two users select a $2\times2$ red square in the lower-left corner and recolor it blue, this represents a shared intention, even if their subsequent actions differ.
This abstraction identifies agreement on what to act on and how, without requiring complete trajectory alignment.

To operationalize this, we extract all (selection, operation) pairs from each trajectory and group them within each task based on spatial and semantic similarity.
This abstracted intention grouping allows us to measure the degree of strategic convergence or diversity across users.
Some tasks exhibit strong convergence, with most participants performing similar key transformations across comparable grid regions.
Others show high diversity, with users selecting different substructures or applying distinct operations.
As shown in Fig.~\ref{fig:compression_histogram}, tasks with low trajectory uniqueness correspond to clearly structured problems with canonical solutions, while those with high uniqueness indicate flexible or ambiguous reasoning spaces.

\textbf{Research Direction 3(a): Strategy Grammar and Reusable Abstractions.}
Future work could formalize intention groupings into a compositional strategy grammar that captures recurring reasoning templates across tasks.
Such a grammar would represent not only sequences of human actions but also higher-level dependencies between subgoals and transformations.
Identifying common strategy motifs, such as ``color and duplicate'' or ``fold and align,'' could support interpretable and human-aligned planning models.
This structured view of recurring human strategies may help models generalize across task categories and explain their reasoning in symbolic or linguistic form.

\textbf{Research Direction 3(b): Intention Prediction and Adaptive Learning.}
Another promising direction is to train models that predict human intention distributions from task features.
By learning how humans allocate attention and plan transformations, such models could estimate which reasoning paths are likely to succeed.
This could guide curriculum sequencing, scaffold reasoning from simpler to diverse tasks, and enable adaptive AI tutors that anticipate user strategies and offer personalized support.
These predictive models could also benchmark AI reasoning diversity, revealing gaps in adaptability and strategic alignment.

\begin{figure*}[htbp]
    \centering
    \includegraphics[width=0.75\textwidth]{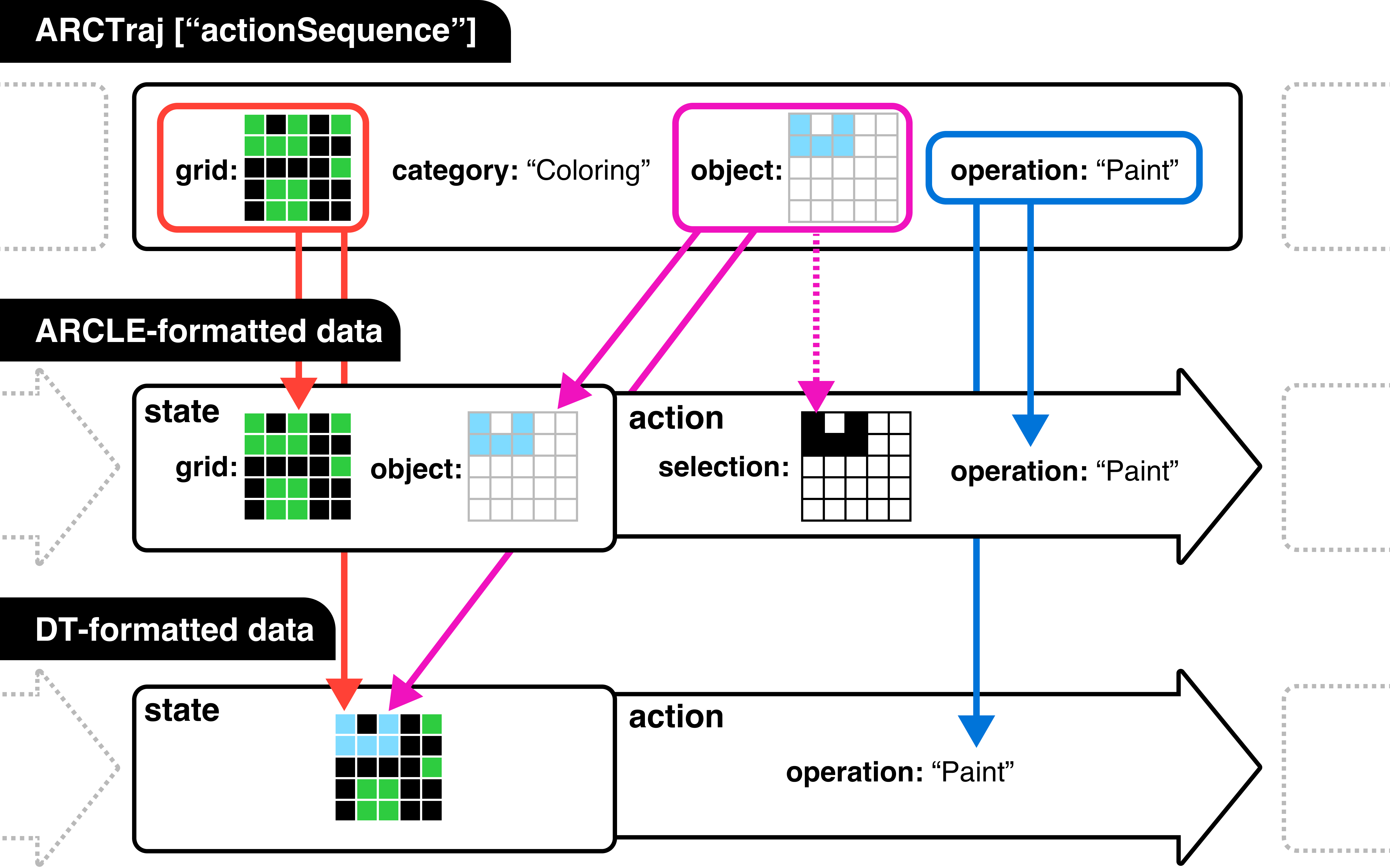}
    \caption{Preprocessing ARCTraj for downstream learning. For RL environments such as ARCLE, ARCTraj is filtered to retain only \texttt{operation} actions and is mapped to a Markovian state-action format using \texttt{grid}, \texttt{object}, and \texttt{operation}. For sequence models such as Decision Transformer, only \texttt{grid} and \texttt{operation} are used, omitting intermediate objects and environment interaction.}
    \Description{Preprocessing ARCTraj for downstream learning. For RL environments such as ARCLE, ARCTraj is filtered to retain only \texttt{operation} actions and is mapped to a Markovian state-action format using \texttt{grid}, \texttt{object}, and \texttt{operation}. For sequence models such as Decision Transformer, only \texttt{grid} and \texttt{operation} are used, omitting intermediate objects and environment interaction.}
    \label{fig:arctraj-preprocess}
\end{figure*}

\section{ARC Solvers using ARCTraj}
\label{sec:solvers}

\subsection{Current ARC Solvers and Key Trends}

Recent research has explored diverse approaches to solving ARC tasks by framing visual reasoning and pattern transformation as learning or generation problems.
As shown in Fig.~\ref{fig:arctraj-preprocess}, ARCTraj data are reformatted for downstream learning across solver paradigms.
For reinforcement learning environments such as ARCLE~\citep{lee2024arcle}, trajectories are mapped to Markovian state–action pairs using \texttt{grid}, \texttt{object}, and \texttt{operation}.
For sequence models such as Decision Transformer~\citep{park2023unraveling}, only \texttt{grid} and \texttt{operation} are retained to simplify intermediate interactions while preserving reasoning context.
This unified preprocessing enables direct comparison of solvers trained on shared human reasoning traces.

Table~\ref{tab:algorithm-performance} summarizes representative models categorized by paradigm, objective, and performance.
These approaches show progress in specific settings but still struggle to generalize reasoning across unseen tasks.
Performance variance suggests that most solvers depend on heuristic priors rather than explicit reasoning mechanisms.

\begin{table*}[htbp]
\centering
\caption{Summary of representative algorithms, categories, and key findings on ARC-related tasks.}
\label{tab:algorithm-performance}
\small
\begin{tabular}{@{}lllll@{}}
\toprule
\textbf{Algorithm / Model} & \textbf{Category} & \textbf{Goal} & \textbf{Performance} & \textbf{Key Findings} \\
\midrule
PPO~\citep{lee2024arcle} & Reinforcement Learning & Solve & 55--70\% & Demonstrated online training in ARC-like MDPs. \\
World Model~\citep{lee2024analogical} & Reinforcement Learning & Solve & 38--100\% & Enabled analogical generalization via latent dynamics. \\
Decision Transformer~\citep{park2023unraveling, kim2025intention} & Sequential Modeling & Solve & 59--90\% & Learned trajectory-conditioned policies with inferred intentions. \\
Diffusion~\citep{kim2024diffusion} & Generative Modeling & Solve & 77--92\% & Generated intermediate states for plan synthesis. \\
GFlowNet~\citep{hwang2025gfn} & Generative Modeling & Augment & 10--100\% & Sampled diverse goal-directed solution trajectories. \\
\bottomrule
\end{tabular}
\end{table*}

Reinforcement learning methods~\citep{lee2024arcle, lee2024analogical} demonstrate online policy learning but require heavy reward shaping and task-specific tuning.
Generative models~\citep{kim2024diffusion, hwang2025gfn} improve output diversity but often rely on statistical associations rather than interpretable abstractions.
Sequential models~\citep{park2023unraveling, kim2025intention} leverage trajectory supervision to mimic human reasoning, but show stable yet limited generalization.
Across paradigms, current solvers still focus on reproducing task outcomes rather than modeling reasoning processes, motivating the use of auxiliary knowledge from ARCTraj as structured cognitive supervision beyond behavioral imitation.

\subsection{Role of ARCTraj in Solver Performance}

ARCTraj contributes to ARC solving by offering auxiliary knowledge that complements traditional training data.
It provides explicit information about human reasoning, including \textit{selection biases}, \textit{color origins}, and \textit{shared intentions}, which can guide model learning.
Integrating these cues helps reduce ambiguity, improve interpretability, and promote generalization across diverse tasks.
In practice, ARCTraj acts as an additional supervision channel that steers learning toward human-like reasoning patterns rather than outcome optimization.

Empirical comparisons show that incorporating ARCTraj signals improves both accuracy and trajectory stability.
For instance, Decision Transformers trained with intention cues gain 6--8 points over demonstration-only baselines.
GFlowNet models augmented with selection priors produce trajectories with 14\% higher diversity and broader coverage of valid transformations.
Diffusion-based solvers integrating intermediate human states achieve more consistent reconstruction of multi-step transformations.
These results indicate that ARCTraj enhances not only outcome quality but also alignment of internal reasoning with human strategies.
Such improvements also enable interpretable evaluation of how models form and revise intermediate hypotheses during task solving.

Despite these gains, ARCTraj remains underutilized.
Most solvers use trajectory data as auxiliary demonstrations rather than structured reasoning supervision.
Further work is needed to formalize how ARCTraj-derived cues interact with solver architectures and to establish principled methods for reasoning-level transfer.

\subsection{Limitations and Research Gaps}

While the models summarized above demonstrate progress, none achieve human-level reasoning on unseen ARC tasks.
The primary limitation is that ARC itself remains an open challenge, requiring compositional generalization and analogical abstraction beyond current learning paradigms.
Most solvers excel at replicating visible transformations but fail to infer implicit rules or relational dependencies.
Moreover, trajectory-based methods have only recently emerged, partly due to the late release of ARCTraj and limited awareness within the research community.
Tooling and benchmarks for trajectory-based reasoning are also in early stages, making systematic evaluation difficult.
Addressing these issues will be essential for advancing toward genuine reasoning-based solutions.

\subsection{Toward Generalizable ARC Solvers}

ARC is inherently challenging because it requires flexible, concept-driven reasoning rather than memorizing patterns.
ARCTraj provides a foundation for addressing this challenge by capturing structured human problem-solving behavior.
Its pipeline, action abstraction, and MDP formulation are transferable to other domains requiring symbolic reasoning and multi-step planning, such as program synthesis, robotic manipulation, and spreadsheet automation.
By viewing ARCTraj not as a static dataset but as a generalizable methodology, future solvers can move toward learning frameworks that model how humans perceive, infer, and act.
This perspective establishes ARCTraj as a bridge between cognitive analysis and machine reasoning, supporting the development of interpretable and adaptable AI systems.

\section{Reproducibility}
\label{sec:reproducibility}

We release ARCTraj and accompanying resources to promote transparency and reproducibility in human-like reasoning research.
The dataset contains over 10,000 human reasoning trajectories collected through the O2ARC platform and structured for compatibility with MDP-based learning.
All data are fully anonymized and publicly accessible through the following repositories:

\begin{itemize}[leftmargin=*]
    \item \textbf{Dataset}: \url{https://huggingface.co/datasets/SejinKimm/ARCTraj}
    \item \textbf{Interactive Viewer}: \url{https://arctraj.github.io/demo}
    \item \textbf{Data Collection Platform}: \url{https://o2arc.com}
\end{itemize}

The O2ARC interface is not open-sourced due to dependency and security constraints, but it provides public access for data collection and analysis.
Notably, the platform features a trajectory recording function that enables users to record and download their own reasoning traces, allowing for further community-driven data collection without exposing the internal backend code.

We also provide open-source implementations of six research projects listed in Table~\ref{tab:algorithm-performance}, covering reinforcement learning, diffusion modeling, sequential reasoning, and intention inference paradigms:

\begin{itemize}[leftmargin=*]
    \item \cite{lee2024arcle}: \url{https://github.com/GIST-DSLab/PPO_Solve}
    \item \cite{lee2024analogical}: \url{https://github.com/GIST-DSLab/Dreamerv3onARCLE}
    \item \cite{kim2025intention}: \url{https://github.com/GIST-DSLab/IntentionLearning}
    \item \cite{park2023unraveling}: \url{https://github.com/GIST-DSLab/ARC_DT}
    \item \cite{kim2024diffusion}: \url{https://github.com/GIST-DSLab/LDCQ}
    \item \cite{hwang2025gfn}: \url{https://github.com/GIST-DSLab/GFN_to_ARC}
\end{itemize}

These repositories include complete training code, preprocessing scripts, and evaluation pipelines, ensuring reproducibility of experiments and facilitating future research built upon ARCTraj.

\section{Conclusion}
\label{sec:conclusion}

ARCTraj captures an extensive collection of human trajectories on ARC tasks, providing fine-grained, step-by-step records of how people engage with abstract visual reasoning problems.
Unlike conventional datasets that include only static input–output pairs, ARCTraj logs temporally ordered, object-level actions grounded in perceptual and symbolic understanding.
This structure provides direct access to intermediate reasoning processes, enabling the detailed modeling of how humans plan, adapt, and transform representations across problem-solving stages.

These structured trajectories have proven useful across multiple learning paradigms.
They enable reinforcement learning agents to be trained from human demonstrations, guide generative planners through trajectory-based augmentation, and support intention-aware models that infer latent subgoals.
The dataset integrates naturally with diverse architectures, including PPO, diffusion models, GFlowNets, and decision transformers, demonstrating its flexibility across imitation- and reasoning-centered research.

Beyond model training, ARCTraj facilitates empirical analysis of cognitive behavior in abstract reasoning.
ARCTraj reveals consistent regularities in attentional selection, color attribution, and strategy convergence, offering evidence of how people decompose complex transformations into subgoals.
These observations bridge cognitive science and AI, providing concrete priors for inductive bias design and evaluation protocols aligned with humans.

Finally, ARCTraj represents more than a dataset; it serves as a methodological framework for linking human reasoning and machine learning.
Its data collection pipeline, trajectory formalism, and MDP transformation can generalize to domains such as program synthesis, robotic manipulation, and real-world planning.
Future work will extend ARCTraj toward larger-scale trajectory collection and adaptive reasoning models that generalize across unseen cognitive tasks.

\section*{Fundings}
This work was supported by IITP (RS-2023-00216011, RS-2024-004450807, No. 2019-0-01842), NRF (RS-2024-00451162, RS-2024-00454000), the InnoCORE program (N10250156), and GIST (Postdoc Value-up) grants funded by the Ministry of Science and ICT, Korea.

\bibliographystyle{ACM-Reference-Format}
\bibliography{99_reference}

\appendix
\appendix
\section{Examples of ARC Tasks and Reasoning}
\label{appendix:arc_task}

The Abstraction and Reasoning Corpus (ARC)~\citep{chollet2019ARC} evaluates abstract reasoning through grid-based transformations.
Each task provides a few demonstrations from which a solver must infer an implicit rule and apply it to a test input.
Solving requires recognizing structural patterns and generalizing from minimal examples without explicit instructions, serving as a benchmark for testing whether models can discover and apply abstract rules from limited evidence.

We illustrate two tasks that highlight distinct reasoning types.
\href{https://o2arc.com/task/c0f76784}{Task
 c0f76784} involves detecting hollow squares and coloring them according to size, reflecting compositional reasoning that combines shape recognition and attribute-based color assignment, showing how ARCTraj captures both discrete and abstract reasoning symbolically.
\href{https://o2arc.com/task/23b5c85d}{Task
 23b5c85d} requires identifying rectangles in the input grid and cropping the one with the smallest area, emphasizing spatial comparison and selection reasoning.

\begin{figure}[htbp]
    \centering
    \includegraphics[width=0.86\linewidth]{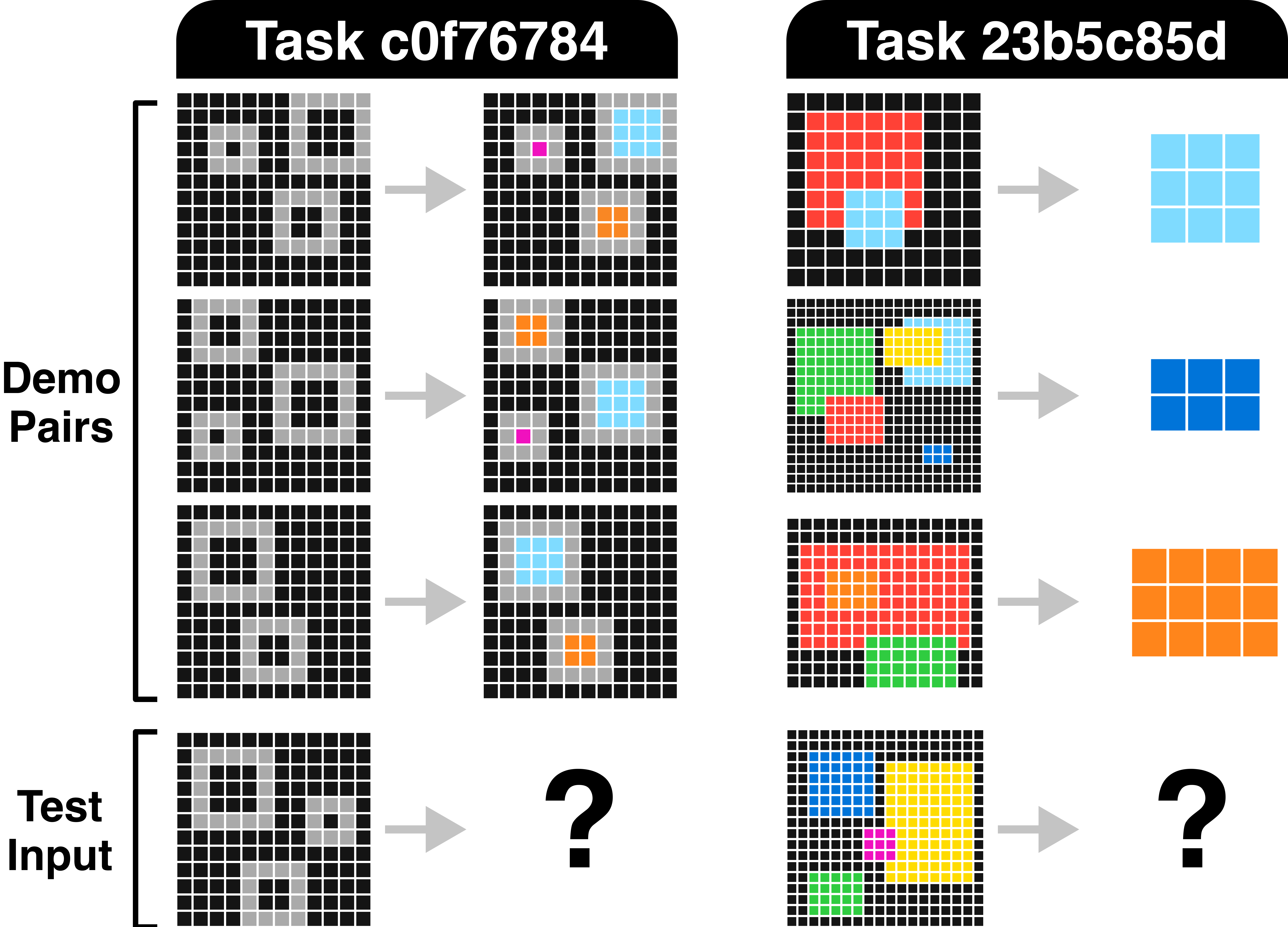}
    \caption{
    Representative ARC tasks used in ARCTraj.
    \href{https://o2arc.com/task/c0f76784}{Task~c0f76784} (left) involves filling hollow squares with color according to their size.
    \href{https://o2arc.com/task/23b5c85d}{Task~23b5c85d} (right) requires selecting the smallest rectangular object and cropping it from the grid.
    These tasks illustrate selection-based and compositional reasoning captured by ARCTraj trajectories.
    }
    \Description{
    Representative ARC tasks used in ARCTraj.
    Task~c0f76784 (left) involves filling hollow squares with color according to their size.
    Task~23b5c85d (right) requires selecting the smallest rectangular object and cropping it from the grid.
    These tasks illustrate selection-based and compositional reasoning captured by ARCTraj trajectories.
    }
    \label{fig:arc_task_examples}
\end{figure}

\section{ARCTraj Schema and Operation Categories}
\label{appendix:o2arc}

Each ARCTraj trajectory is stored as a JSON object describing grid operations.
Each record logs the category, operation type, coordinates, and local grid snapshot for interpretable reconstruction of human reasoning.
This compact representation abstracts user interactions into a structured format for analysis and reasoning.
Listing~\ref{lst:colored_json} shows an example of a single recorded action in this schema.

\lstdefinelanguage{ColoredJSON}{
  basicstyle=\ttfamily\footnotesize,
  showstringspaces=false,
  breaklines=true,
  frame=single,
  backgroundcolor=\color{gray!5},
  numberstyle=\tiny,
  numbers=none,
  stringstyle=\color{black}, 
  morestring=[b]",
  literate=
   *{0}{{{\color{black}0}}}{1}
    {1}{{{\color{blue}1}}}{1}
    {3}{{{\color{green!70!black}3}}}{1}
    {4}{{{\color{orange!85!black}4}}}{1}
    {6}{{{\color{magenta}6}}}{1}
    {:}{{{\color{black}:}}}{1}
    {,}{{{\color{black},}}}{1}
}

\begin{lstlisting}[
  language=ColoredJSON,
  caption={Example JSON data from the ``actionSequence'' column of ARCTraj, logging the operation type, position, grid state, and selected object. (Field names ‘x,y’ are data keys.)},
  label={lst:colored_json}
]
{
  "category": "Selection",
  "operation": "SelectCell",
  "position": {
    "x": 7,
    "y": 9
  },
  "grid": [
    [0,0,0,0,0,0,0,0,0,0,0,0,0,0,0,0,0,0],
    [0,0,0,0,0,0,0,0,0,0,0,0,0,0,0,0,0,0],
    [0,0,1,1,1,1,1,1,0,0,0,0,0,0,0,0,0,0],
    [0,0,1,1,1,1,1,1,0,4,4,4,4,4,4,4,4,0],
    [0,0,1,1,1,1,1,1,0,4,4,4,4,4,4,4,4,0],
    [0,0,1,1,1,1,1,1,0,4,4,4,4,4,4,4,4,0],
    [0,0,1,1,1,1,1,1,0,4,4,4,4,4,4,4,4,0],
    [0,0,1,1,1,1,1,1,0,4,4,4,4,4,4,4,4,0],
    [0,0,0,0,0,0,0,0,0,4,4,4,4,4,4,4,4,0],
    [0,0,0,0,0,0,0,6,6,6,4,4,4,4,4,4,4,0],
    [0,0,0,0,0,0,0,6,6,6,4,4,4,4,4,4,4,0],
    [0,0,0,0,0,0,0,6,6,6,4,4,4,4,4,4,4,0],
    [0,0,0,0,0,0,0,0,0,4,4,4,4,4,4,4,4,0],
    [0,0,3,3,3,3,3,0,0,4,4,4,4,4,4,4,4,0],
    [0,0,3,3,3,3,3,0,0,0,0,0,0,0,0,0,0,0],
    [0,0,3,3,3,3,3,0,0,0,0,0,0,0,0,0,0,0],
    [0,0,3,3,3,3,3,0,0,0,0,0,0,0,0,0,0,0],
    [0,0,0,0,0,0,0,0,0,0,0,0,0,0,0,0,0,0]
  ],
  "object": [
    {
      "x": 7,
      "y": 9,
      "color": 6
    }
  ],
  "overlapped": true,
  "timestamp": "2024-02-15T01:07:40.537Z"
}
\end{lstlisting}

Table~\ref{tab:arctraj_ops} summarizes the symbolic operation categories identified in ARCTraj and their representative operation types.
These categories organize low-level interface actions into higher-level abstractions that describe how humans manipulate and reason about grid transformations.

\begin{table}[htbp]
    \centering
    \caption{
    Summary of operation categories and representative operation types in ARCTraj.
    Each category groups related symbolic actions extracted from user trajectories.
    }
    \begin{tabular}{@{}ll@{}}
        \toprule
        \textbf{Category} & \textbf{Operations} \\
        \midrule
        Selection & \texttt{SelectCell}, \texttt{SelectGrid}, \texttt{SelectObject} \\
        Coloring & \texttt{Paint} \\
        Critical & \texttt{ResizeGrid}, \texttt{Submit} \\
        O2 & \texttt{Flip}, \texttt{Move}, \texttt{Rotate} \\
        History & \texttt{Redo}, \texttt{Undo} \\
        Clipboard & \texttt{Copy}, \texttt{Paste} \\
        \bottomrule
    \end{tabular}
    \label{tab:arctraj_ops}
\end{table}
\section{Analytic Extensions of ARCTraj}
\label{appendix:analysis}

We formalize three analytic modules introduced in the main paper, \textit{Selection Bias} (Sec.~\ref{sec:selection_biases}), \textit{Color Origins} (Sec.~\ref{sec:color_origin}), and \textit{Shared Intentions} (Sec.~\ref{sec:shared_intention}), which quantify how human reasoning trajectories reflect exploration, perceptual grounding, and convergent strategies in ARC task solving.
Each analysis translates behavioral regularities into measurable variables for systematic comparison across tasks.

\subsection{Selection Bias in Exploration}
\label{appendix:selection}

Human solvers do not always follow goal-directed attention; they often begin by exploring uncertain or salient regions before identifying the correct transformation.
This exploratory phase shows how people generate and refine hypotheses, reflecting the balance between perceptual search and goal-oriented reasoning in ARC tasks.
To quantify this tendency, we measure how attention diverges from solution-relevant regions.

We define a spatial bias metric comparing the empirical selection distribution \(p_{\text{sel}}(i, j)\) from ARCTraj and the object-region distribution \(p_{\text{obj}}(i, j)\):
\[
\text{Bias} = \mathrm{KL}\!\left(p_{\text{sel}} \,\|\, p_{\text{obj}}\right),
\]
where \(\mathrm{KL}\) denotes the Kullback–Leibler divergence.
A higher bias indicates dispersed, exploratory selections that deviate from relevant regions, while a lower bias reflects focused attention aligned with solution objects.
Tracking bias over time also shows how exploration narrows as solvers converge on the rule.

The bias decomposes into two components:
(i) \textit{dispersion}, how widely selections spread across the grid, and
(ii) \textit{misalignment}, how often attention targets irrelevant areas.
The first reflects perceptual uncertainty, the second conceptual exploration.
These distinguish tasks that cause visual confusion from those that require abstract reasoning.
Dispersion can be measured by spatial entropy, and the overlap ratio quantifies the degree of misalignment.

Selection also evolves over time.
Let each selection at time \(t\) be \(S_t = (i_t, j_t, w_t, h_t)\), producing a selection sequence \(\mathcal{S} = \{S_1, \ldots, S_T\}\).
A latent-phase model,
\[
P(\text{phase}_t \mid \mathcal{S}_{1:t-1}), \quad \text{where } \text{phase}_t \in \{\texttt{explore}, \texttt{exploit}\},
\]
captures transitions between exploratory and exploitative reasoning.
Early exploration shows high-entropy selections scattered across the grid, while later exploitation yields localized, repetitive selections.
Entropy decay in \(\mathcal{S}\) provides a quantitative signal of reasoning convergence, aligning with decreasing spatial bias.

These analyses apply directly to few-shot solvers \(f_\theta\).
Bias measures can serve as auxiliary supervision, guiding the solver to refine attention or reasoning schedules.
Regularizing attention maps to penalize dispersion or reward phase-consistent focus allows \(f_\theta\) to emulate human exploratory balance.
In observed tasks, ambiguous or repetitive structures show higher mean bias and inter-user variance, consistent with uncertainty-driven exploration.

Together, the spatial and temporal selection patterns define auxiliary knowledge
\[
\mathcal{A}_{\text{selection}} =
\big\{
p_{\text{sel}}(i, j),\,
\text{Bias}(p_{\text{sel}}, p_{\text{obj}}),\,
P(\text{phase}_t \mid \mathcal{S}_{1:t-1})
\big\},
\]
which provides spatial and temporal priors for \(f_\theta\).

\begin{figure*}[htbp]
    \centering
    \begin{subfigure}[b]{0.24\textwidth}
        \centering
        \includegraphics[width=\textwidth]{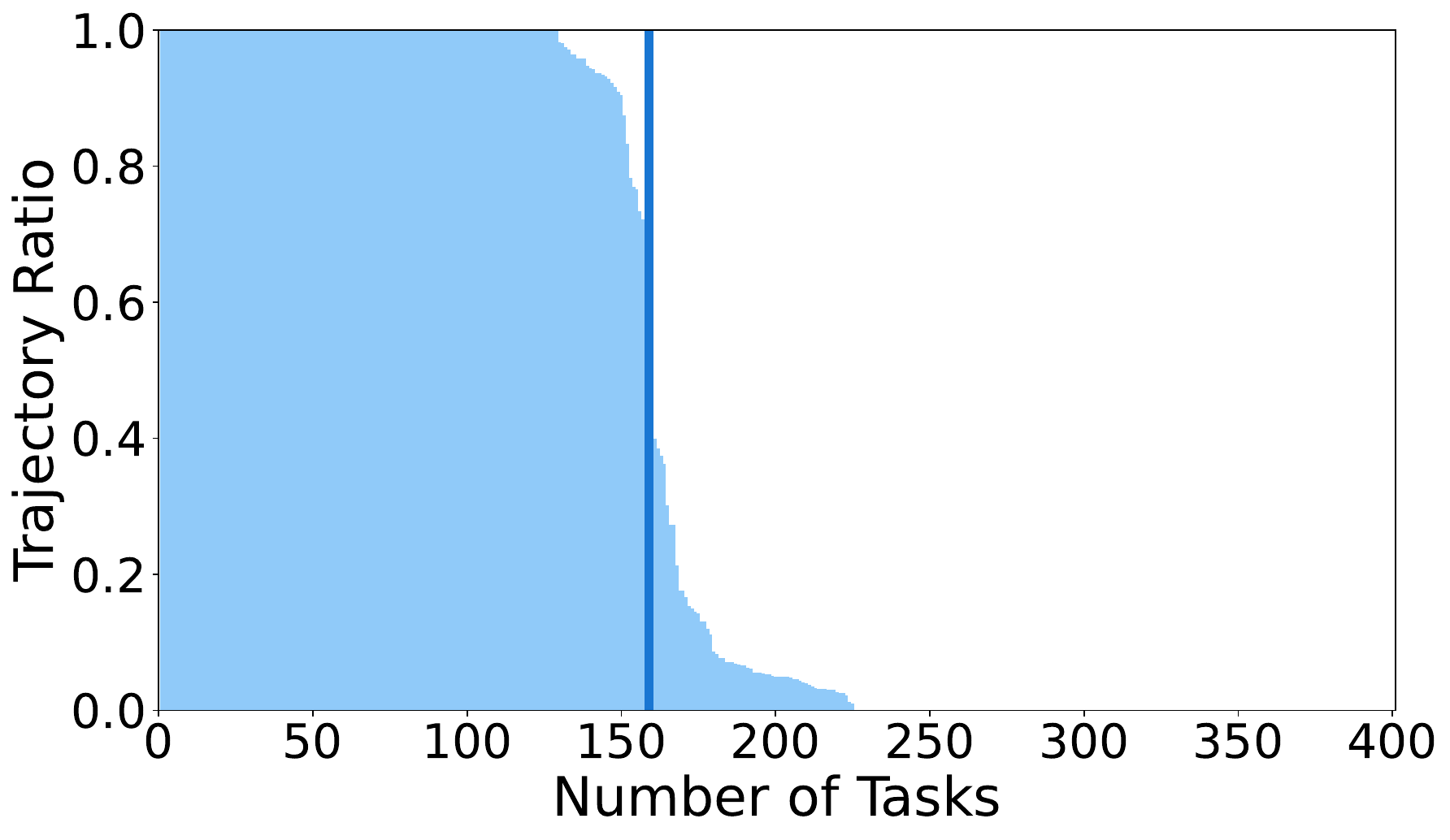}
        \caption{$\mathcal{C}^{\text{demo},\text{I}}$}
        \label{fig:color_origin_demoI}
    \end{subfigure}
    \hfill
    \begin{subfigure}[b]{0.24\textwidth}
        \centering
        \includegraphics[width=\textwidth]{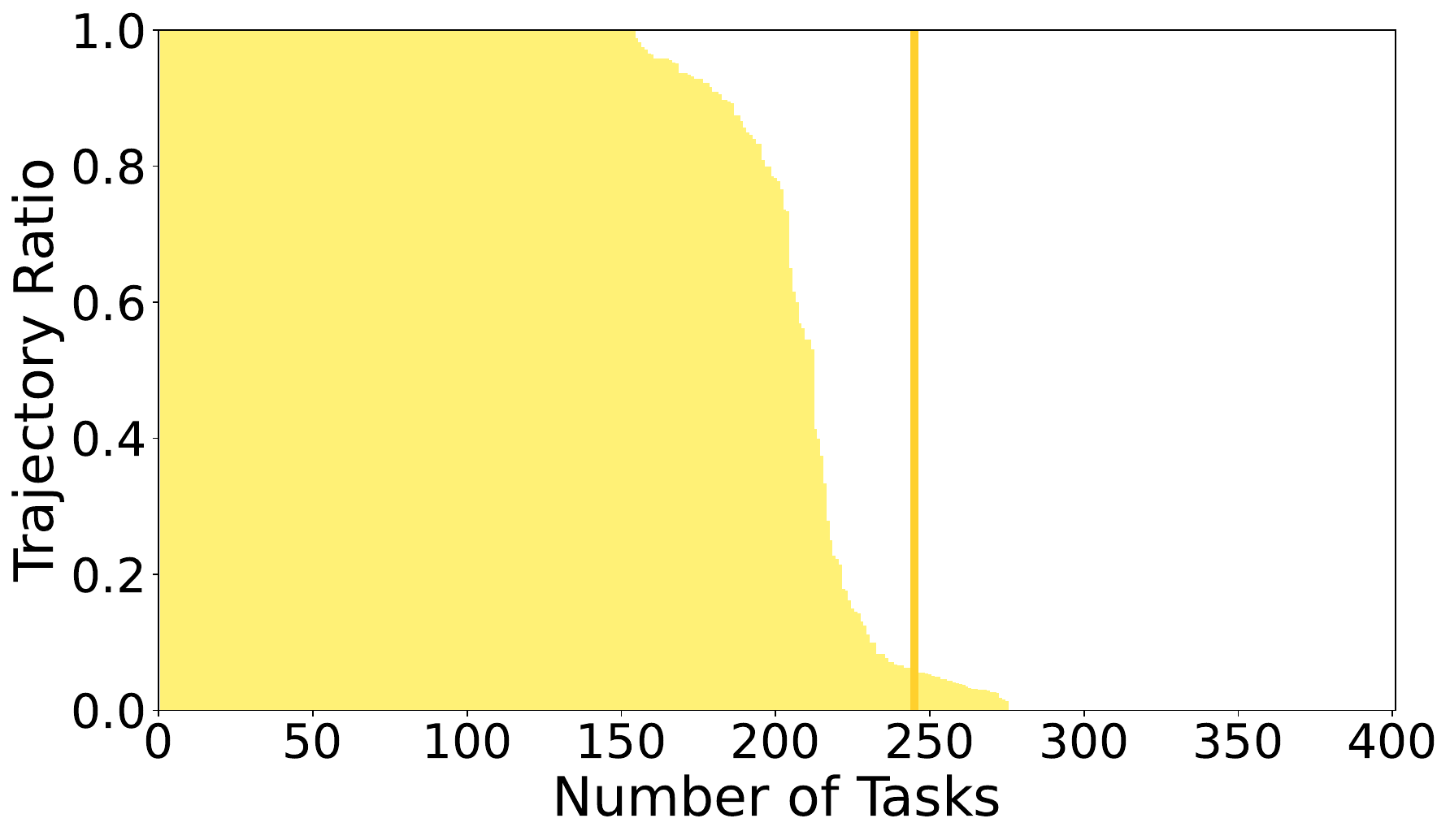}
        \caption{$\mathcal{C}^{\text{demo},\text{O}}$}
        \label{fig:color_origin_demoO}
    \end{subfigure}
    \hfill
    \begin{subfigure}[b]{0.24\textwidth}
        \centering
        \includegraphics[width=\textwidth]{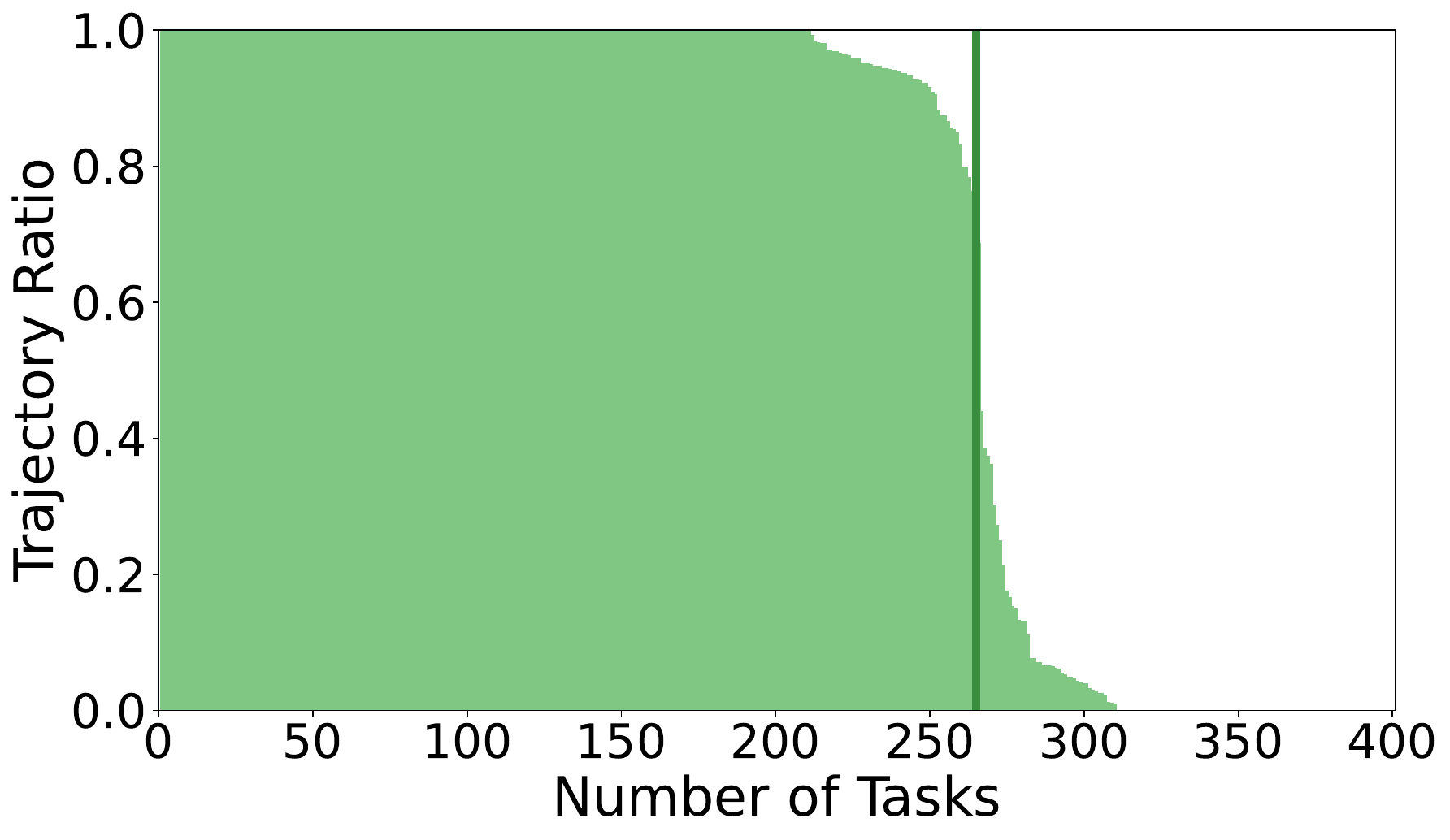}
        \caption{$\mathcal{C}^{\text{test},\text{I}}$}
        \label{fig:color_origin_testI}
    \end{subfigure}
    \hfill
    \begin{subfigure}[b]{0.24\textwidth}
        \centering
        \includegraphics[width=\textwidth]{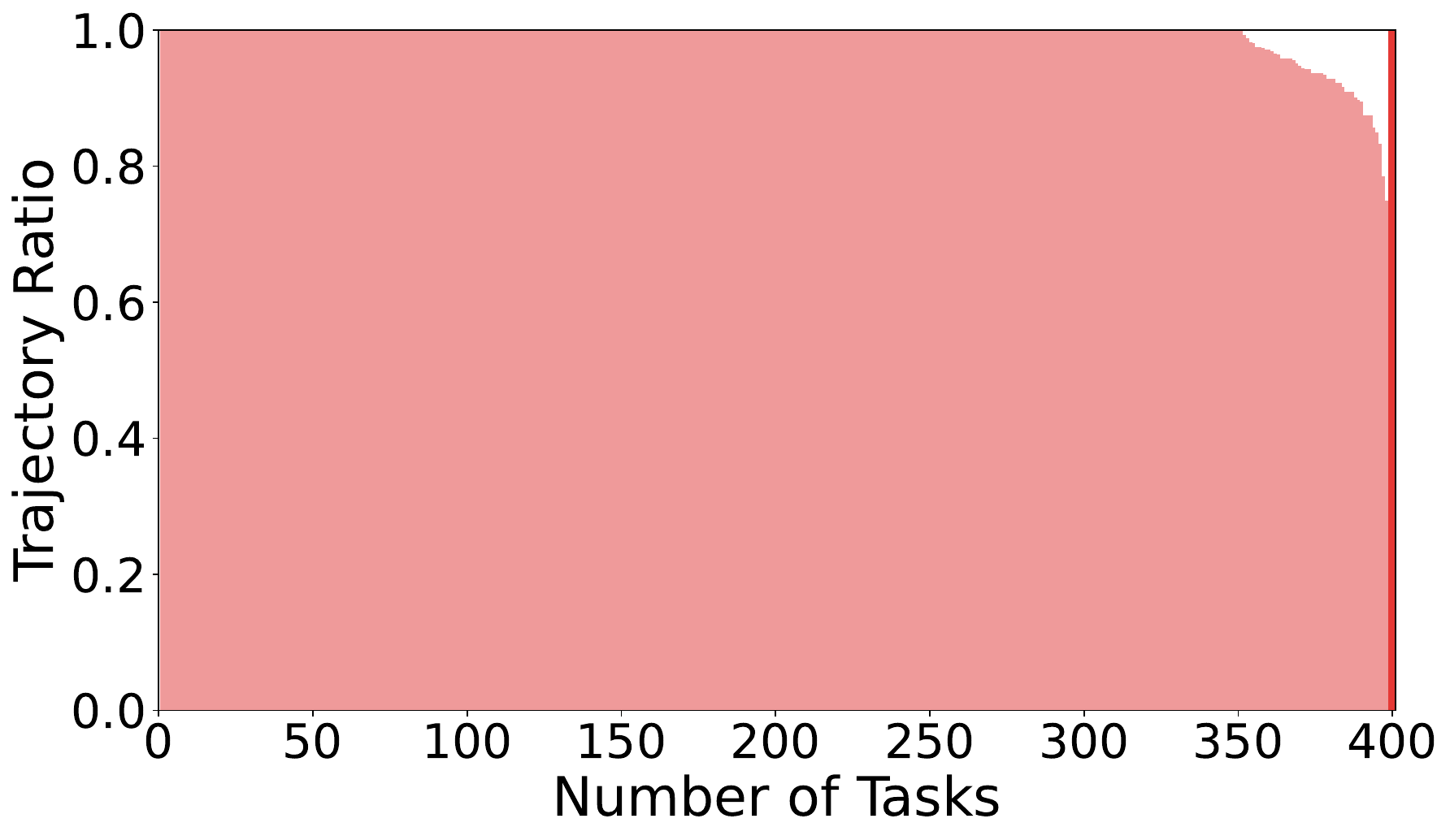}
        \caption{$\mathcal{C}^{\text{union}}$}
        \label{fig:color_origin_union}
    \end{subfigure}
    
    \caption{
    Distribution of trajectory ratios across 400 ARC tasks, showing discrepancies between theoretical color requirements and actual human usage.
    Each subplot displays the proportion of trajectories using colors from a specific source (\(\mathcal{C}^{\text{demo},\text{I}}\), \(\mathcal{C}^{\text{demo},\text{O}}\), \(\mathcal{C}^{\text{test},\text{I}}\), \(\mathcal{C}^{\text{union}}\)), with the vertical line marking how many tasks require that source.
    In \(\mathcal{C}^{\text{demo},\text{I}}\), example input colors are often underused despite being theoretically needed, whereas \(\mathcal{C}^{\text{demo},\text{O}}\) reveals frequent overuse of demonstration output colors.
    \(\mathcal{C}^{\text{test},\text{I}}\) aligns closely with task requirements, suggesting a human bias toward test inputs.
    \(\mathcal{C}^{\text{union}}\) confirms that all tasks are solvable using only in-task colors, consistent with ARC’s constrained design.
    These trends highlight inductive biases that extend beyond the availability of colors, motivating the explicit modeling of color origins.
    }
    \Description{
    Distribution of trajectory ratios across 400 ARC tasks, showing discrepancies between theoretical color requirements and actual human usage.
    Each subplot displays the proportion of trajectories using colors from a specific source (\(\mathcal{C}^{\text{demo},\text{I}}\), \(\mathcal{C}^{\text{demo},\text{O}}\), \(\mathcal{C}^{\text{test},\text{I}}\), \(\mathcal{C}^{\text{union}}\)), with the vertical line marking how many tasks require that source.
    In \(\mathcal{C}^{\text{demo},\text{I}}\), example input colors are often underused despite being theoretically needed, whereas \(\mathcal{C}^{\text{demo},\text{O}}\) reveals frequent overuse of demonstration output colors.
    \(\mathcal{C}^{\text{test},\text{I}}\) aligns closely with task requirements, suggesting a human bias toward test inputs.
    \(\mathcal{C}^{\text{union}}\) confirms that all tasks are solvable using only in-task colors, consistent with ARC’s constrained design.
    These trends highlight inductive biases that extend beyond the availability of colors, motivating the explicit modeling of color origins.
    }
    \label{fig:color_origin}
\end{figure*}

\subsection{Color Abstraction and Origin Analysis}
\label{appendix:colorset}

Human solvers often reuse or reinterpret colors beyond those shown in demonstrations.
Their color choices reveal how people abstract visual features and infer symbolic relations across grid regions.
Analyzing these behaviors provides auxiliary knowledge \(\mathcal{A}\) that helps solvers generalize color reasoning across tasks.

We formalize color provenance through four candidate sources:
colors from example inputs \(\mathcal{C}^{\text{demo},\text{I}}\), example outputs \(\mathcal{C}^{\text{demo},\text{O}}\), test inputs \(\mathcal{C}^{\text{test},\text{I}}\), and their union \(\mathcal{C}^{\text{union}}\).
Each \texttt{paint} action is represented as \(a_t = (\texttt{paint}, c_t, S_t)\) where \(c_t \in \{0,\ldots,9\}\) is the target color and \(S_t =(i_t, j_t, w_t, h_t)\) the selected region.
A source function
\[
\text{src}(c_t) \in \{\mathcal{C}^{\text{demo},\text{I}}, \mathcal{C}^{\text{demo},\text{O}}, \mathcal{C}^{\text{test},\text{I}}\}
\]
maps each color to its probable origin, yielding a conditional model
\[
P\big(\text{src}(c_t)\mid \mathcal{D}_{\text{demo}}, x^{\text{test}}, \mathcal{A}\big),
\]
where contextual features include local color frequencies and spatial proximity to colored regions.
This captures how solvers associate new colors with known references, forming a probabilistic model of color transfer.

Beyond literal provenance, humans generalize colors by semantic function.
We define an abstraction mapping
\[
\psi : S_t \rightarrow \text{abstract color role},
\]
where \(\psi\) groups regions into roles such as “object,” “background,” or “mirror fill.”
Learning \(\psi\) through clustering or manual labeling provides a structured bridge from perceptual to symbolic reasoning, extending color inference beyond visual matching.

Together, provenance estimation and functional abstraction define auxiliary knowledge
\[
\mathcal{A}_{\text{color}} = \{\text{src}(c_t), \psi(S_t)\}.
\]
The resulting conditional model for color prediction is
\[
P\big(c_t \mid \mathcal{D}_{\text{demo}}, x^{\text{test}}, \mathcal{A}_{\text{color}} \big),
\]
which provides perceptual and semantic priors for \( f_\theta \).
Tasks with low color-entropy distributions typically involve perceptually grounded transformations, whereas higher entropy reflects symbolic reinterpretation and conceptual generalization.

\subsection{Shared Intentions Across Participants}
\label{appendix:intention}

Human solvers often reach similar intermediate goals despite taking different actions.
These shared intentions are recurring combinations of spatial selections and symbolic operations that serve as reusable reasoning units.
They form mid-level structures that link local perception with global task goals, demonstrating how people reuse procedural knowledge across tasks.

Each intention is defined as a pair \( I_t = (S_t, \text{op}_t) \),
where \(S_t\) is the selected region and \(\text{op}_t\) the applied operation (e.g., \texttt{Paint}, \texttt{Move}).
Given two trajectories \( \mathcal{I}_a \) and \( \mathcal{I}_b \), their similarity is
\[
\text{Sim}(\mathcal{I}_a, \mathcal{I}_b) = 
\frac{\lvert I_a \cap I_b \rvert}{\lvert I_a \cup I_b \rvert},
\]
where each \(I\) denotes a unique \((S, \text{op})\) pair normalized for scale and rotation.
Higher similarity indicates that different solvers rediscover similar task decompositions.

Aggregating trajectory similarities yields a population matrix identifying frequent intention clusters.
Each cluster corresponds to a reasoning template such as “select–paint,” “copy–align,” or “mirror–extend.”
These clusters represent procedural patterns that recur across tasks, showing that human reasoning reuses a limited set of compositional motifs.
Formally,
\[
\mathcal{M} = \{ M_k \}_{k=1}^K, \quad 
M_k = \{ (S_t, \text{op}_t) \mid (S_t, \text{op}_t) \in \text{cluster } k \}.
\]
The number of clusters \(K\) depends on task diversity but remains small relative to all trajectories, suggesting that reasoning relies on modular reuse rather than random exploration.

Shared-intention clusters provide interpretable priors for constructing compositional reasoning in few-shot solvers \( f_\theta \).
Aligning solver states with human-derived prototypes offers structured guidance on how to apply operations within a trajectory.

Empirically, tasks with high inter-user similarity show faster convergence and lower selection entropy, reflecting alignment around stable subgoals.
Tasks with low similarity involve ambiguous or compositional transformations, where solvers follow diverse yet coherent subgoals.
These findings show that shared intentions link perception-driven exploration with symbolic planning, forming auxiliary knowledge \(\mathcal{A}_{\text{intention}} = \mathcal{M}\) that contributes to the overall reasoning prior \(\mathcal{A}\) used by \( f_\theta \).

\begin{figure*}[htbp]
    \centering
    \includegraphics[width=0.88\linewidth]{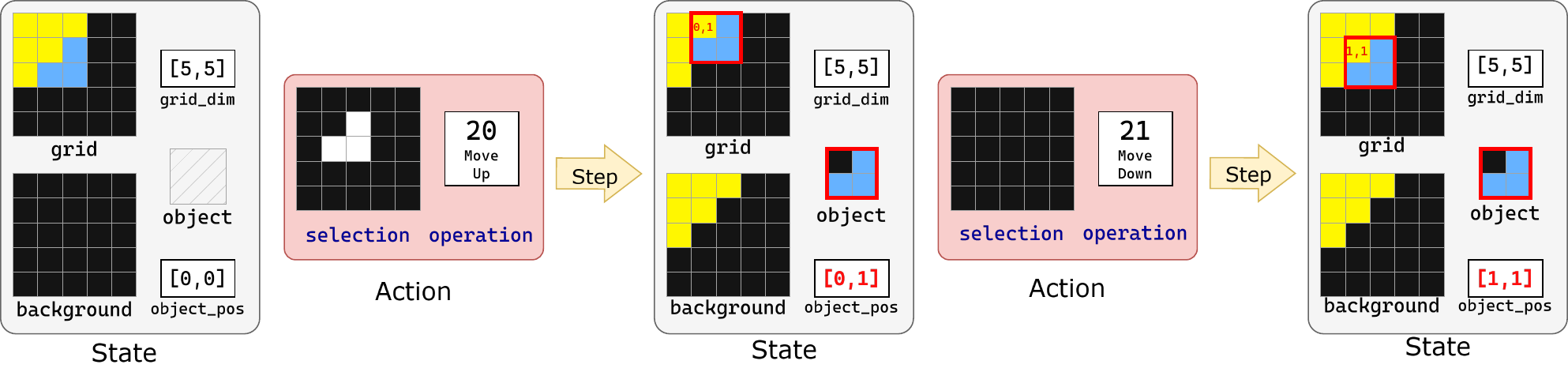}
    \caption{State transition in ARCLE~\citep{lee2024arcle}: the agent observes the current grid and applies a symbolic editing action to update it. This process models ARC as a Markov decision problem, where the agent sequentially edits the grid to reach a correct solution.}
    \Description{State transition in ARCLE: the agent observes the current grid and applies a symbolic editing action to update it. This process models ARC as a Markov decision problem, where the agent sequentially edits the grid to reach a correct solution.}
    \label{fig:arcle_pipeline}
\end{figure*}

\section{Use of ARCTraj in Learning-Based Solvers}
\label{appendix:use_arctraj}

\subsection{ARCLE and RL Framework}

ARCLE~\citep{lee2024arcle} formulates the ARC challenge as a reinforcement learning (RL) problem, where the output grid acts as the environment and symbolic editing operations define the discrete action space. This framing enables end-to-end learning of sequential decision-making policies directly over symbolic grid transformations.

A key difficulty is the sparsity of rewards: the agent receives a reward only when producing an entirely correct final output. Such sparse signals hinder standard RL training, as most trajectories yield no reward. To alleviate this, ARCLE leverages ARCTraj’s structured human demonstrations for behavior cloning, which guides the policy toward promising regions of the ample search space.

ARCLE uses ARCTraj’s \texttt{actionSequence} to learn a mapping from states to expert actions, and then fine-tunes the policy with PPO to generalize beyond demonstrations and handle unseen states. Follow-up work~\citep{lee2024analogical} integrates DreamerV3 with ARCTraj-based initialization, further improving sample efficiency and planning.

Fig.~\ref{fig:arcle_pipeline} illustrates ARCLE’s state–action transitions within an MDP framework. The agent observes grid states and selects symbolic edits that move the grid toward the target solution. Empirical results show that ARCTraj-based imitation substantially accelerates convergence and stabilizes training compared to pure online RL.

ARCLE also uses ARCTraj to train a World Model that predicts future states and actions in a latent space. This model captures temporal structure, supports multi-step planning, and improves task generalization, further linking human problem-solving with autonomous ARC agent learning.

\subsection{World Model Training with ARCTraj Data}

Building on the RL framework, ARCLE incorporates a DreamerV3-based World Model~\citep{hafner2023dreamerv3} to learn compact latent representations of ARC tasks from the richly annotated ARCTraj data. This latent modeling improves planning and policy learning by mapping high-dimensional grid and symbolic action states into a continuous, lower-dimensional space that preserves essential task features.

The architecture comprises encoder--decoder pairs for grid states and symbolic actions, a recurrent hidden state $h_t$ that models temporal dependencies, and a reconstruction-based objective. Training on ARCTraj’s structured trajectories enables the model to predict future latent states and actions, supporting lookahead reasoning and more informed decision-making.

\begin{figure}[htbp]
    \centering
    \includegraphics[width=\linewidth]{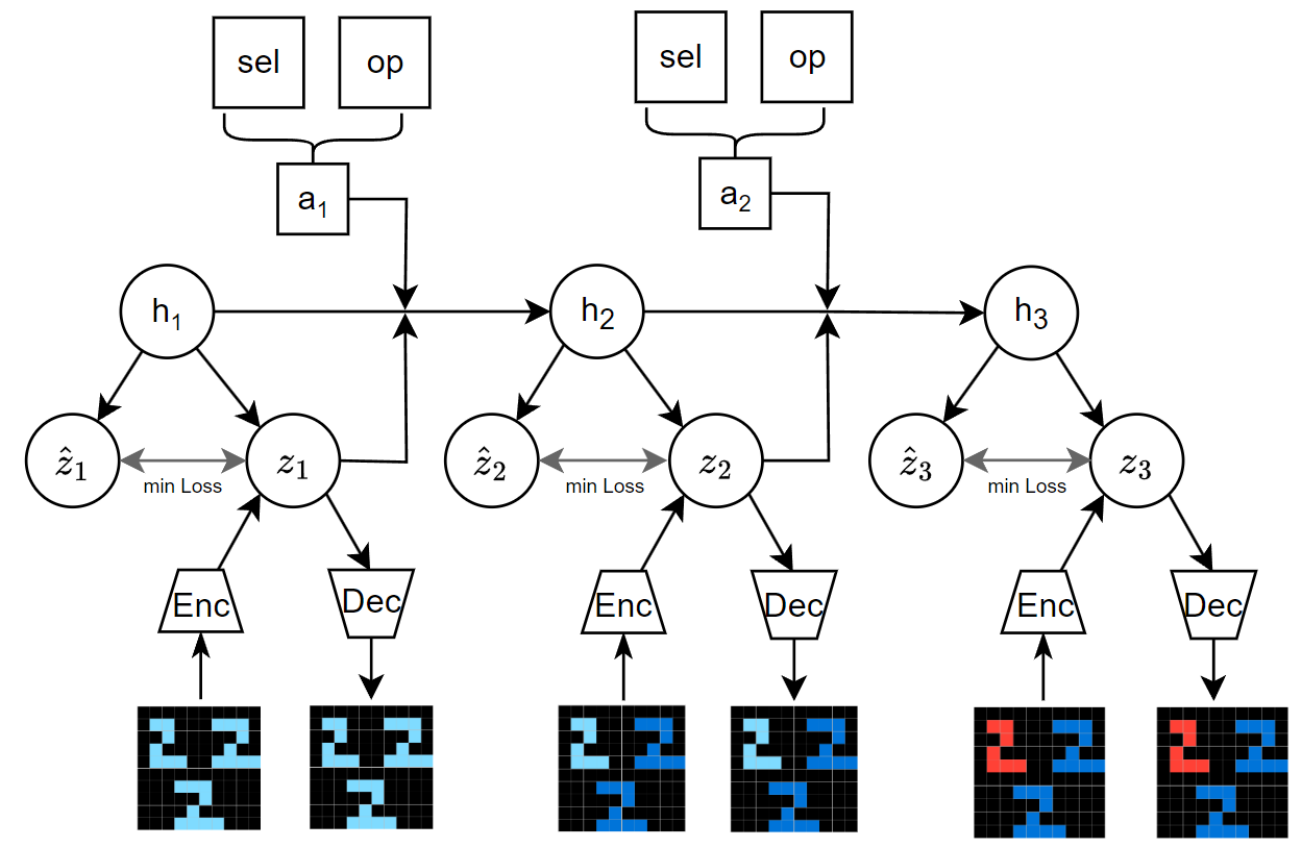}
    \caption{The DreamerV3-based World Model~\citep{lee2024analogical} is trained on ARCTraj using an encoder–decoder architecture that embeds grid states and symbolic operations into latent variables $z_t$. A recurrent state $h_t$ captures temporal dependencies between latents and actions $a_t$. Minimizing reconstruction loss enables accurate recovery of states and operations. Using ARCTraj’s rich trajectories, the model predicts future states and actions in a compact latent space, improving planning, sample efficiency, and generalization on ARC tasks.}
    \Description{The DreamerV3-based World Model~\citep{lee2024analogical} is trained on ARCTraj and uses an encoder--decoder architecture to embed grid states and symbolic operations into compact latents $z_t$. A recurrent state $h_t$ tracks temporal dependencies between latents and actions $a_t$, capturing task dynamics. By minimizing reconstruction loss, the model accurately recovers states and operations from the latent space. Leveraging ARCTraj’s detailed trajectories, it predicts future states and actions for planning in a low-dimensional space, improving sample efficiency and generalization on complex ARC tasks.}
    \label{fig:arcle_world_model}
\end{figure}

Fig.~\ref{fig:arcle_world_model} illustrates the overall structure. Grid and action states are encoded into latent variables \(z_t\), providing compact summaries of visual configurations and symbolic transformations. The recurrent state \(h_t\) aggregates temporal information, enabling the model to track multi-step dependencies. Decoders reconstruct the original states from the latents, and the training loss encourages accurate prediction of grid transitions and action effects. This design models both spatial and temporal structure, supporting latent rollouts that align with trajectories observed in ARCTraj.

Leveraging ARCTraj’s detailed symbolic trajectories enables the World Model to capture meaningful environmental dynamics and action semantics specific to ARC tasks. This latent dynamics modeling improves sample efficiency by enabling planning in a compact space rather than the whole grid. Empirical results demonstrate that ARCTraj-trained World Models enhance policy learning, accelerate convergence, and improve generalization on unseen ARC tasks.
\begin{figure*}[htbp]
    \centering
    \includegraphics[width=0.9\linewidth]{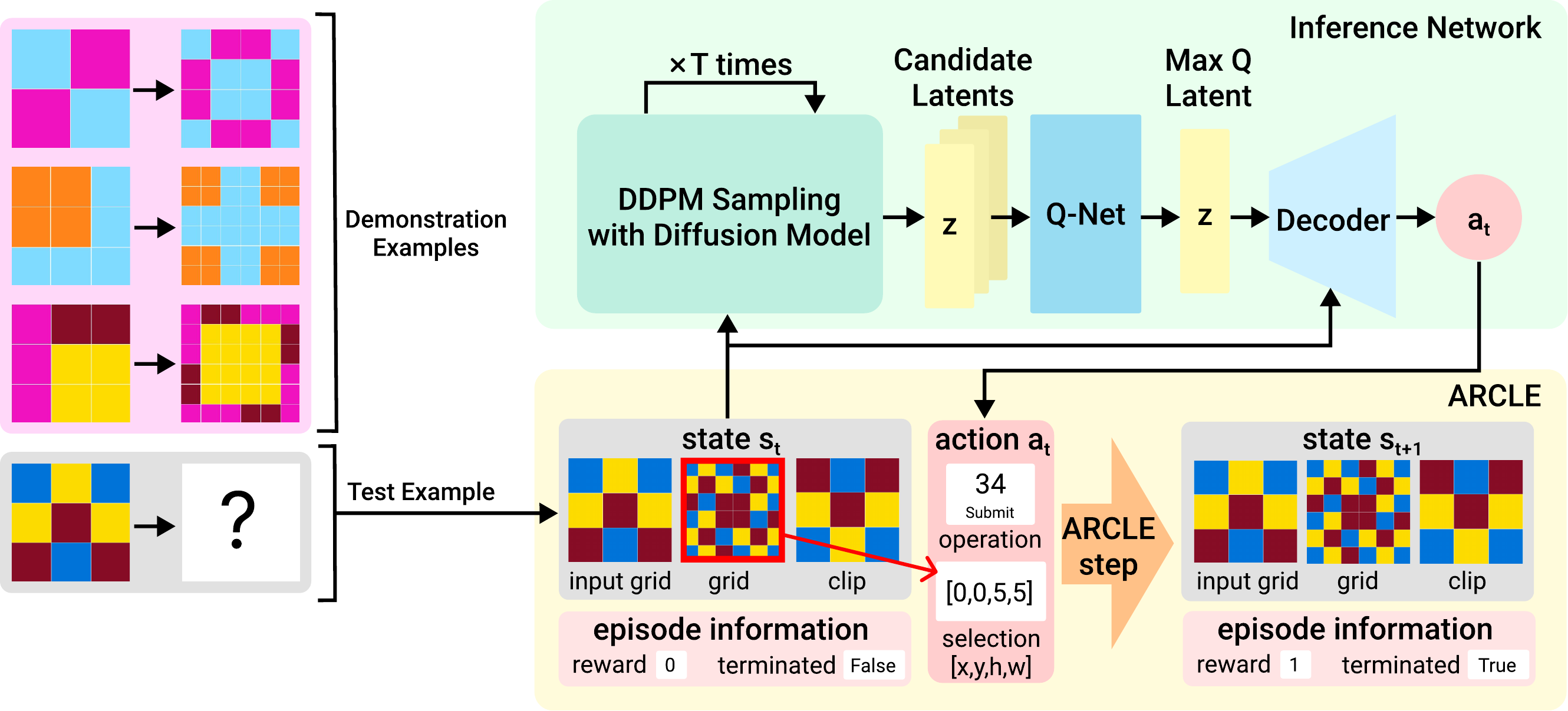}
    \caption{LDCQ trains an autoencoder on ARCTraj’s symbolic action sequences to learn a continuous latent space for discrete operations~\citep{kim2024diffusion}. The encoder embeds action trajectories into latent representations, and the decoder reconstructs actions from them. Constrained Q-learning then optimizes a policy within this latent space, using rewards to guide exploration and ensure valid solutions. This continuous space enables smoother diffusion-based sampling, thereby reducing the challenges associated with large discrete action spaces. Grounding the latent space in human trajectories from ARCTraj further improves sample efficiency and solution quality.}
    \Description{LDCQ trains an autoencoder on ARCTraj’s symbolic action sequences to learn a continuous latent space for discrete operations~\citep{kim2024diffusion}. The encoder embeds action trajectories into latent representations, and the decoder reconstructs actions from them. Constrained Q-learning then optimizes a policy within this latent space, using rewards to guide exploration and ensure valid solutions. This continuous space enables smoother diffusion-based sampling, thereby reducing the challenges associated with large discrete action spaces. Grounding the latent space in human trajectories from ARCTraj further improves sample efficiency and solution quality.}
    \label{fig:ldcq_arch}
\end{figure*}

\subsection{Diffusion-based Offline RL}

LDCQ (Latent Diffusion with Constrained Q-Learning)~\citep{kim2024diffusion} introduces an approach that integrates latent diffusion models with constrained Q-learning to address the compositional and multi-step nature of ARC tasks. Instead of operating directly in a discrete symbolic action space, where optimization becomes difficult due to sparsity and discontinuity, LDCQ constructs a continuous latent action space using an autoencoder trained on ARCTraj's rich symbolic trajectories. This latent space captures semantic relationships among symbolic operations, placing functionally similar actions closer together and enabling smoother transitions during policy learning. By learning such structure-aware embeddings, the model gains the ability to reason over symbolic operations more flexibly and expressively.

With this latent representation, LDCQ employs diffusion-based generative modeling to explore actions through gradual denoising, guided by learned score functions. This continuous exploration mitigates the combinatorial explosion of discrete operations, enabling the model to generalize beyond the exact operation sequences observed in the dataset. Moreover, the diffusion process provides a powerful mechanism for interpolating between symbolic actions, enabling the policy to propose candidate edits that blend characteristics of multiple operations in a coherent way. Constrained Q-learning then steers the diffusion sampling toward high-reward and structurally valid regions of the latent space. The Q-function incorporates ARC-specific constraints, ensuring that sampled latent vectors correspond to plausible editing steps and stable intermediate grid states, ultimately improving the reliability of predicted transformations and enhancing overall robustness.

ARCTraj plays a central role in shaping this latent action space. Human-generated trajectories encode strong inductive biases, such as local reasoning, symmetry exploitation, and multi-step compositional edits, that become reflected in the learned embeddings. Because the latent representations arise from real human strategies, they naturally encode structural regularities that benefit downstream decision-making and help the model understand how symbolic edits are typically sequenced and combined in realistic problem-solving scenarios. This grounding enables the model to generate action sequences with meaningful symbolic semantics rather than arbitrary latent encodings. Also, it helps preserve coherence across successive edits by providing latent features that implicitly capture task context. As a result, both policy expressiveness and interpretability are improved, since the generated transformations more closely resemble the reasoning patterns humans exhibit when working through multi-step ARC manipulations.

By combining diffusion-based exploration with constraint-aware Q-learning, LDCQ effectively navigates the ample combinatorial solution space of ARC. The synergy between a structured latent action manifold and reward-guided sampling yields policies that explore more broadly while maintaining alignment with feasible symbolic reasoning. Empirical results show higher success rates and faster convergence compared to traditional RL methods, owing to smoother optimization dynamics and better alignment with human reasoning patterns in ARCTraj. Overall, LDCQ demonstrates how integrating continuous generative modeling with human-derived symbolic data can significantly enhance AI reasoning capabilities in abstract problem-solving domains, indicating a promising direction for hybrid symbolic and continuous learning frameworks.

\subsection{GFlowNet-based Trajectory Augmentation}

The GFlowNet-based approach~\citep{hwang2025gfn} frames ARC task solving as a structured sequence generation problem, where each trajectory is a series of symbolic editing operations that transform the input grid into the target output. Instead of optimizing for a single best solution as in traditional reinforcement learning, GFlowNet learns a generative policy that samples trajectories with probabilities proportional to their rewards. This probabilistic formulation facilitates the discovery of diverse, high-quality solutions that more accurately reflect the variety of human problem-solving strategies.

A central challenge is the combinatorial explosion of possible trajectories, which makes naive exploration infeasible. ARCTraj alleviates this by supplying human-generated trajectories that act as high-reward exemplars. These demonstrations guide the model’s sampling distribution toward promising regions of the solution space, helping it acquire effective policies more efficiently.

Training incorporates these augmented trajectories into a flow matching objective that balances exploration and exploitation, enabling GFlowNet to represent multiple modes in the solution space. Leveraging ARCTraj reduces the likelihood of the model collapsing into narrow local optima and helps it capture the underlying structure of ARC tasks. This leads to improved sample efficiency, robustness, and more plausible solutions than those generated by standard RL agents.

\begin{figure}[htbp]
    \centering
    \includegraphics[width=\linewidth]{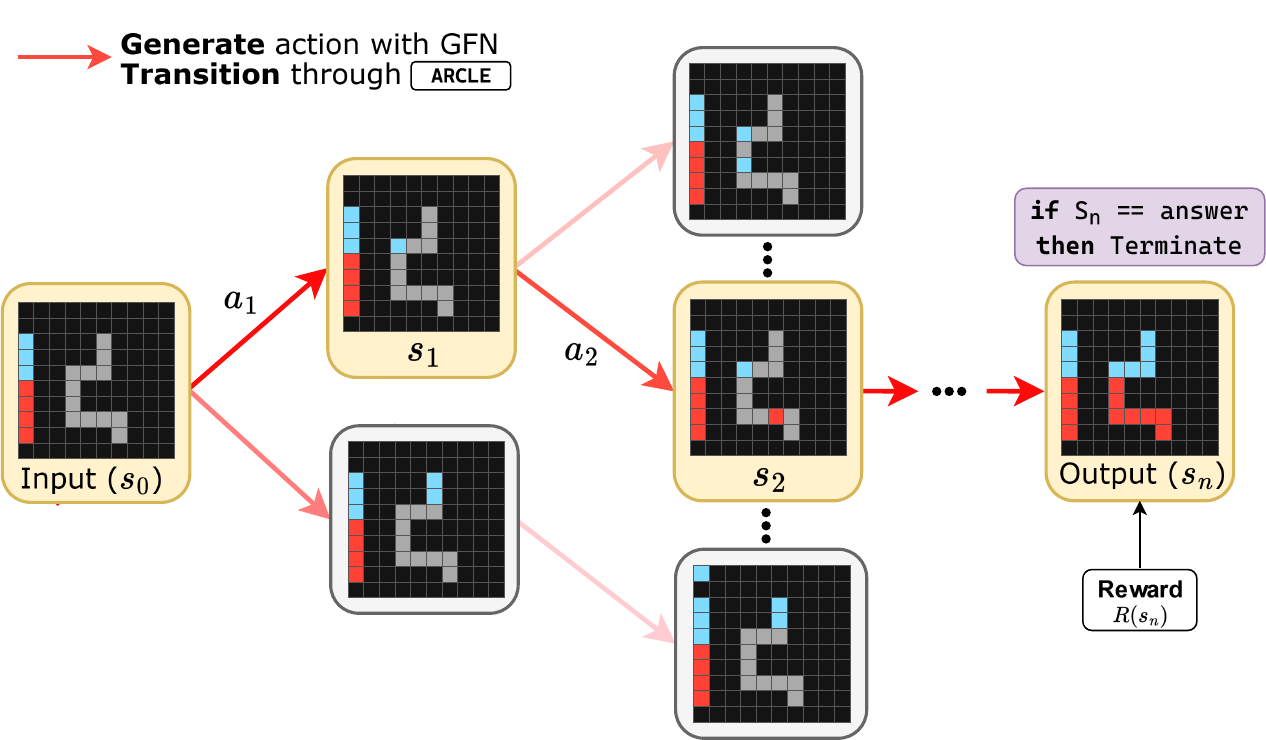}
    \caption{Overview of the GFlowNet-based ARC solver that leverages ARCTraj’s human trajectories as reward-supporting paths~\citep{hwang2025gfn}. The model generates symbolic editing sequences with probabilities proportional to their rewards, and the human trajectories guide flow matching toward high-quality solution paths. This probabilistic training promotes diverse exploration and efficient learning across ARC’s complex, multi-modal solution landscape.}
    \Description{Overview of the GFlowNet-based ARC solver that leverages ARCTraj’s human trajectories as reward-supporting paths~\citep{hwang2025gfn}. The model generates symbolic editing sequences with probabilities proportional to their rewards, and the human trajectories guide flow matching toward high-quality solution paths. This probabilistic training promotes diverse exploration and efficient learning across ARC’s complex, multi-modal solution landscape.}
    \label{fig:gfn_aug}
\end{figure}

In summary, incorporating ARCTraj data into GFlowNet training improves the solver’s ability to navigate the complex solution space, allowing it to produce diverse and practical strategies. This probabilistic, data-augmented approach enhances learning efficiency and enables the model to capture the structure and variety inherent in ARC tasks. Consequently, it represents a promising direction for developing AI agents with more human-like reasoning and adaptability, capable of tackling challenges that demand creativity and exploration beyond traditional optimization methods.

\subsection{Decision Transformer}

Recent work~\citep{park2023unraveling} frames ARC task solving as a sequential decision-making problem and applies Decision Transformers (DTs) to model user behavior from offline data. DTs treat the task as sequence modeling, predicting the next action conditioned on the current grid state, past actions, and the expected future returns (\textit{return-to-go}). By converting ARCTraj’s symbolic trajectories into \texttt{(return-to-go, state, action)} triples, DTs learn human problem-solving strategies directly from offline logs. Our work extends this by adding object-level information, forming \texttt{(return-to-go, state, action, object)} quadruplets to better capture spatial reasoning.

The DT architecture uses a transformer to model long-range temporal dependencies in trajectories. Conditioning on return-to-go guides the model toward trajectories associated with higher rewards, improving its ability to imitate effective solution patterns and predict human-aligned next actions.

\begin{figure}[htbp]
    \centering
    \includegraphics[width=\linewidth]{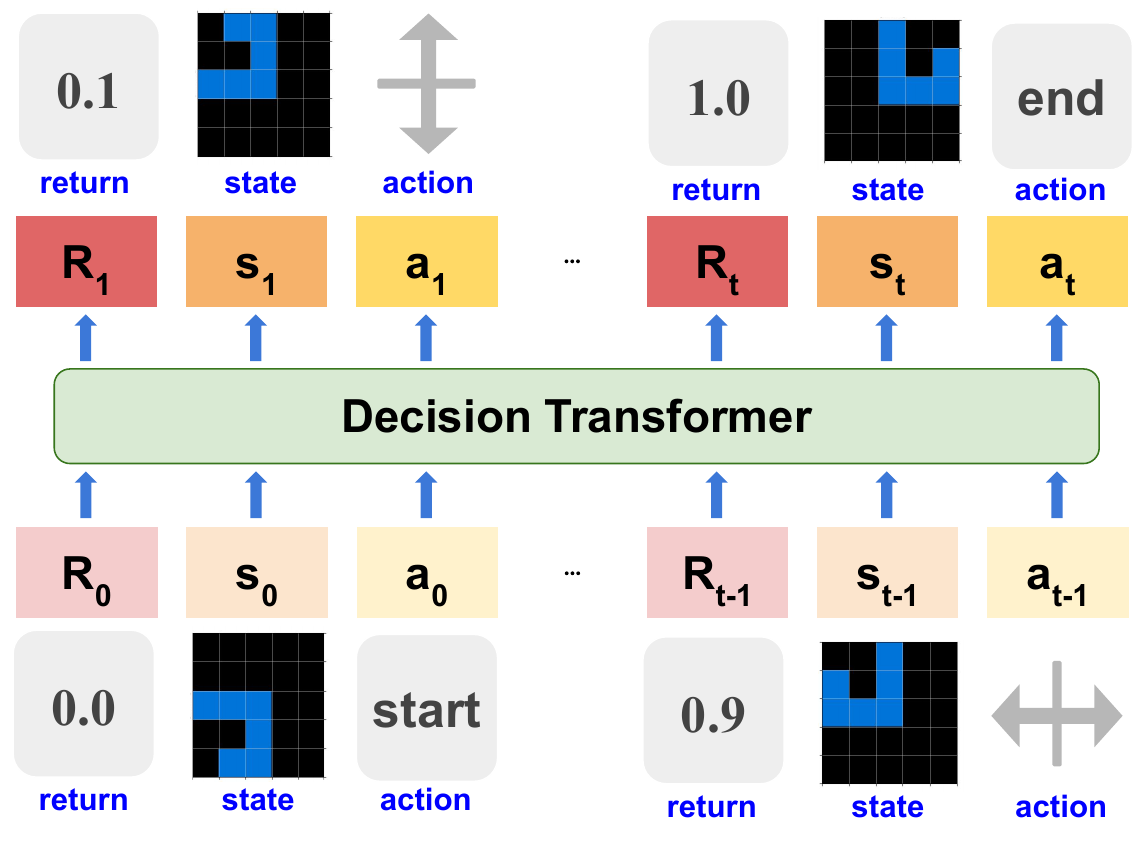}
    \caption{Decision Transformer-based ARC solver trained on ARCTraj’s symbolic action sequences and return-to-go signals~\citep{park2023unraveling}. The transformer receives sequences of past returns, states, actions, and optionally object or intention embeddings, enabling it to predict next editing operations and replicate human-like reasoning. This offline framework leverages rich human trajectories to improve policy learning without costly environment rollouts.}
    \Description{Decision Transformer-based ARC solver trained on ARCTraj’s symbolic action sequences and return-to-go signals~\citep{park2023unraveling}. The transformer receives sequences of past returns, states, actions, and optionally object or intention embeddings, enabling it to predict next editing operations and replicate human-like reasoning. This offline framework leverages rich human trajectories to improve policy learning without costly environment rollouts.}
    \label{fig:dt_training}
\end{figure}

Building on this framework, recent work~\citep{kim2025intention} incorporates high-level intention embeddings from user logs, forming pentaplet samples \texttt{(return-to-go, state, action, object, intention)}. This additional conditioning enables the DT to model more abstract planning behaviors and produce coherent, higher-level action sequences that generalize better to unseen ARC tasks.

Together, these approaches demonstrate the effectiveness of pairing advanced sequence modeling with rich human trajectory data. The Decision Transformer framework supports scalable offline training, paving the way for interpretable ARC agents that incorporate cognitive structures, such as intentions and hierarchical planning in a consistent manner.

\section{Exploratory Data Analysis}


\subsection{Comparison with H-ARC Dataset}

We compare ARCTraj with H-ARC~\citep{legris2024harc}, a dataset of human editing sequences on ARC tasks. H-ARC logs pixel-level edits, where each action merges selection and color change. ARCTraj instead captures object-level sequences via the O2ARC interface~\citep{shim2024o2arc}, which separates selection from symbolic operations such as \texttt{Move}, \texttt{Rotate}, \texttt{Flip}, \texttt{Copy}, and \texttt{Paste}. This object-centric design reveals higher-level structural intentions that pixel-level logs cannot capture.

Although both datasets target the same tasks, their formats lead to different strategy patterns: H-ARC reflects pixel-level edits, while ARCTraj highlights semantic transformations over shapes and relations. These differences influence trajectory length and action composition in meaningful ways. Sec.~\ref{sec:dataset-statistics} shows the high-level comparison, and the following sections extend it with detailed quantitative measurements for deeper analysis.

\subsection{Quantitative Comparison Statistics}

Statistics from the Training Sets of both datasets are computed using the same metrics presented in Table~\ref{tab:arctraj_basic} and Table~\ref{tab:arctraj_vs_harc}. All ARCTraj values employ the same methodology to ensure direct comparability. Beyond the raw counts, we also highlight how each metric relates to reasoning behavior and the types of cognitive patterns it reveals, offering additional interpretive depth.

\begin{table}[ht]
\setlength{\abovecaptionskip}{-5pt}
\centering
\begin{tabular}{lrr}
\toprule
\textbf{Metric} & \textbf{ARCTraj} & \textbf{H-ARC} \\
\midrule
Average participants per task & 13.9 & 11.8 \\
Average trajectories per task & 25.5 & 19.8 \\
Number of trajectories & 10,193 & 7,916 \\
Ratio of object level actions & 15.2\% (37.7\%) & 0.9\% \\
Number of actions & 208,721 (84,123) & 241,697 \\
Number of object level actions & 31,710 & 2,227 \\
Ratio of cross trajectory grids & 43.7\% & 11.4\% \\
Number of cross trajectory grids & 14,688 & 7,834 \\
Number of unique grids & 33,608 & 68,914 \\
\bottomrule
\end{tabular}
\caption{
Extended comparison between ARCTraj and H-ARC. This table aggregates the statistics underlying Table~\ref{tab:arctraj_basic} and Table~\ref{tab:arctraj_vs_harc}, providing a compact summary of the EDA results.
}
\label{tab:arctraj_harc_comparison}
\end{table}

\paragraph{Average Participants per Task}
ARCTraj records an average of 13.9 unique participants per task, compared to 11.8 in H-ARC. The slightly higher value indicates broader coverage of per-user tasks. This is relevant because users who attempt more tasks often exhibit recognizable cross-task reasoning habits, such as repeatedly applying structural decompositions or symmetry-based strategies across different problem families.

\paragraph{Average Trajectories per Task}
ARCTraj contains 10,193 valid trajectories for the 400 training tasks (25.5 per task), whereas H-ARC contains 7,916 trajectories (19.8 per task). Higher trajectory density in ARCTraj yields more stable estimates of strategy variability and intermediate-state distributions. It also supports fine-grained clustering analyses, where trajectory groups can be compared based on the degree of abstraction they employ.

\paragraph{Number of Trajectories}
ARCTraj includes 10,193 trajectories, compared to 7,916 in H-ARC. This 29\% increase expands coverage of rare reasoning strategies and reduces sensitivity to participant-specific biases, yielding more representative aggregate patterns.

\paragraph{Ratio of Object-level Actions}
Object-level actions are detected when \texttt{category} is \texttt{O2} or \texttt{Clipboard}. In ARCTraj, these constitute 31,710 out of 208,721 actions (15.2\%) when all selection steps are included, or 31,710 out of 84,123 (37.7\%) when selections are excluded. In H-ARC, the ratio is 0.9\%. This large gap clearly illustrates how O2ARC enables users to perform conceptual transformations directly, rather than approximating them through extensive pixel changes in manual editing.

\paragraph{Number of Actions}
Depending on the treatment of selection steps, ARCTraj trajectories contain 208,721 actions (raw), 137,152 actions (merged selections), or 84,123 actions (operations only). H-ARC logs 241,697 pixel-level actions. ARCTraj trajectories remain shorter even under the least compressed measurement. This reduction reflects conceptual compression: a single \texttt{Rotate} action in ARCTraj corresponds to many pixel-level operations in H-ARC.

\paragraph{Number of Object-level Actions}
ARCTraj logs 31,710 object-level edits, while H-ARC logs 2,227. This fourteen-fold difference highlights the degree to which ARCTraj captures geometric and relational patterns that require many operations to express in pixel-level logs. These higher-level operations provide richer supervision signals for models aiming to learn structured transformations.

\paragraph{Ratio of Cross-trajectory Grids}
ARCTraj contains 14,688 shared grids out of 33,608 unique states (43.7\%), while H-ARC contains 7,834 shared states out of 68,914 (11.4\%). The higher convergence rate in ARCTraj suggests that object-level reasoning guides solvers toward more consistent intermediate states. This phenomenon is useful for inferring latent strategy templates and for studying how humans organize multi-step transformations.

\paragraph{Number of Unique Grids}
ARCTraj records 33,608 unique states, while H-ARC records 68,914. Despite ARCTraj’s larger trajectory volume, the reduced number of unique states indicates that object-centric actions compress the state space by focusing on structurally meaningful transitions. This compression is advantageous for downstream modeling because it reduces spurious variability.

\subsection{Key Observations}

The extended statistics reinforce the differences outlined in Section~\ref{sec:dataset-statistics}. 
H-ARC is suited for pixel-level analysis, whereas ARCTraj more directly captures high-level conceptual transformations enabled by object-centric operations. 
ARCTraj exhibits higher trajectory density, significantly larger ratios of object-level actions (15.2 to 37.7\%), and shorter action sequences, indicating that users rely on compact symbolic edits rather than long pixel-level chains.

ARCTraj also shows stronger cross-trajectory convergence (43.7\% versus 11.4\%), suggesting that solvers frequently reach similar intermediate configurations despite varying edit paths. This consistency highlights recurrent transformation patterns and stable subgoals, 
providing useful supervision signals for modeling human-like reasoning. 
A consolidated summary of these numerical results is presented in Table~\ref{tab:arctraj_harc_comparison}, offering a clear overview of the overall trends.

\end{document}